\documentclass[10pt,twocolumn,letterpaper]{article}

\usepackage{cvpr}
\usepackage{times}
\usepackage{epsfig}
\usepackage{graphicx}
\usepackage{amsmath}
\usepackage{amssymb}
\usepackage{dblfloatfix}
\usepackage{nicefrac}
\usepackage{subcaption}
\usepackage{array}
\usepackage{adjustbox}
\usepackage[cc]{titlepic}

\newcolumntype{x}[1]{>{\centering\arraybackslash\hspace{0pt}}p{#1}}

\newlength{\myheightA}
\newlength{\myheightB}
\usepackage[pagebackref=true,breaklinks=true,letterpaper=true,colorlinks,bookmarks=false]{hyperref}

\cvprfinalcopy 

\def\cvprPaperID{2104}

\ifcvprfinal\fi
\begin{document}

\title{Efficient Object Localization Using Convolutional Networks}

\author{Jonathan Tompson, Ross Goroshin, Arjun Jain, Yann LeCun, Christoph Bregler\\
New York University\\
{\tt\small tompson/goroshin/ajain/lecun/bregler@cims.nyu.edu}
}

\twocolumn[{%
\renewcommand\twocolumn[1][]{#1}%
\maketitle
\begin{center}
    \centering
\adjustbox{height=\myheightA}
      {\includegraphics[width=\textwidth]{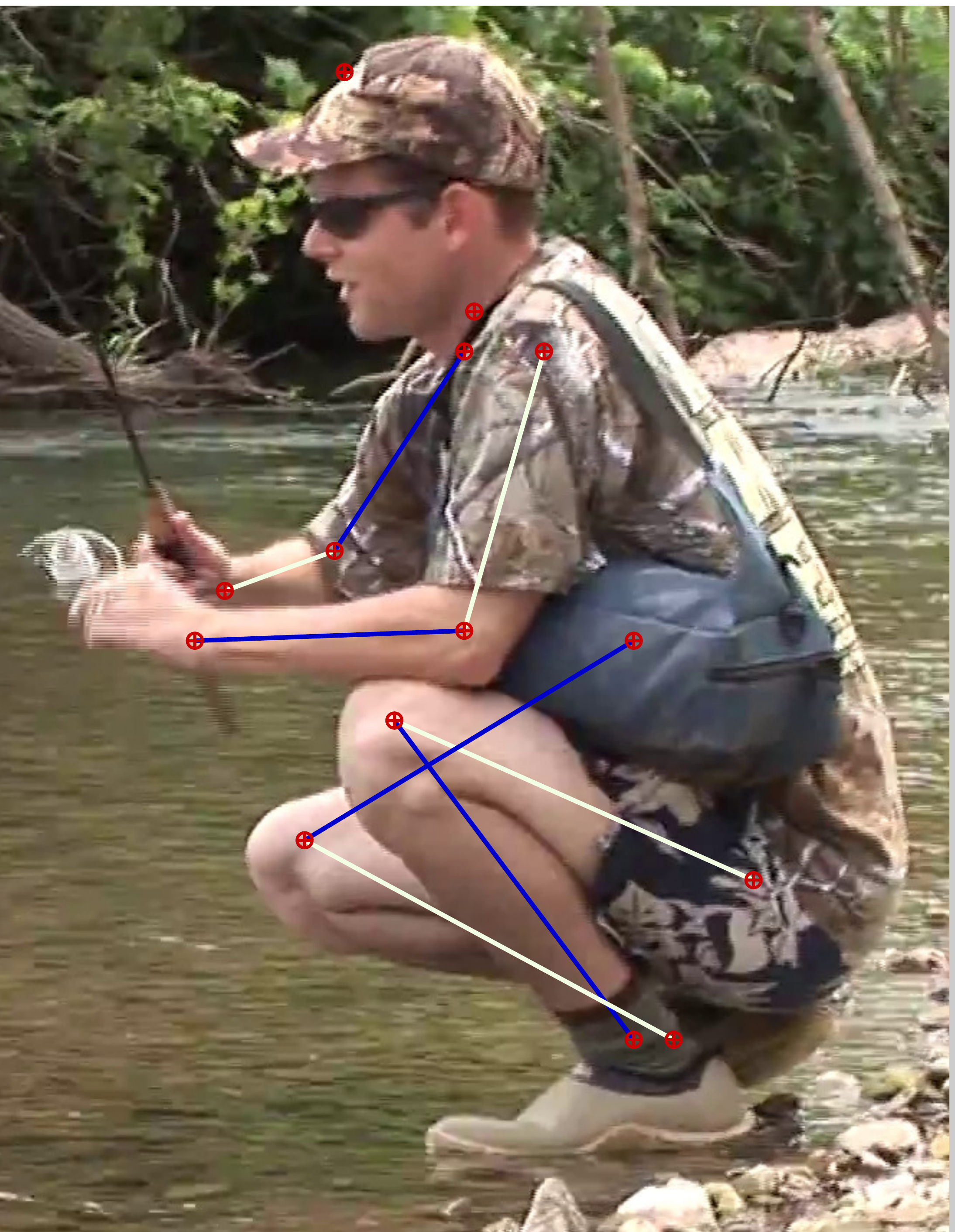}}
  \adjustbox{height=\myheightA}
      {\includegraphics[width=\textwidth]{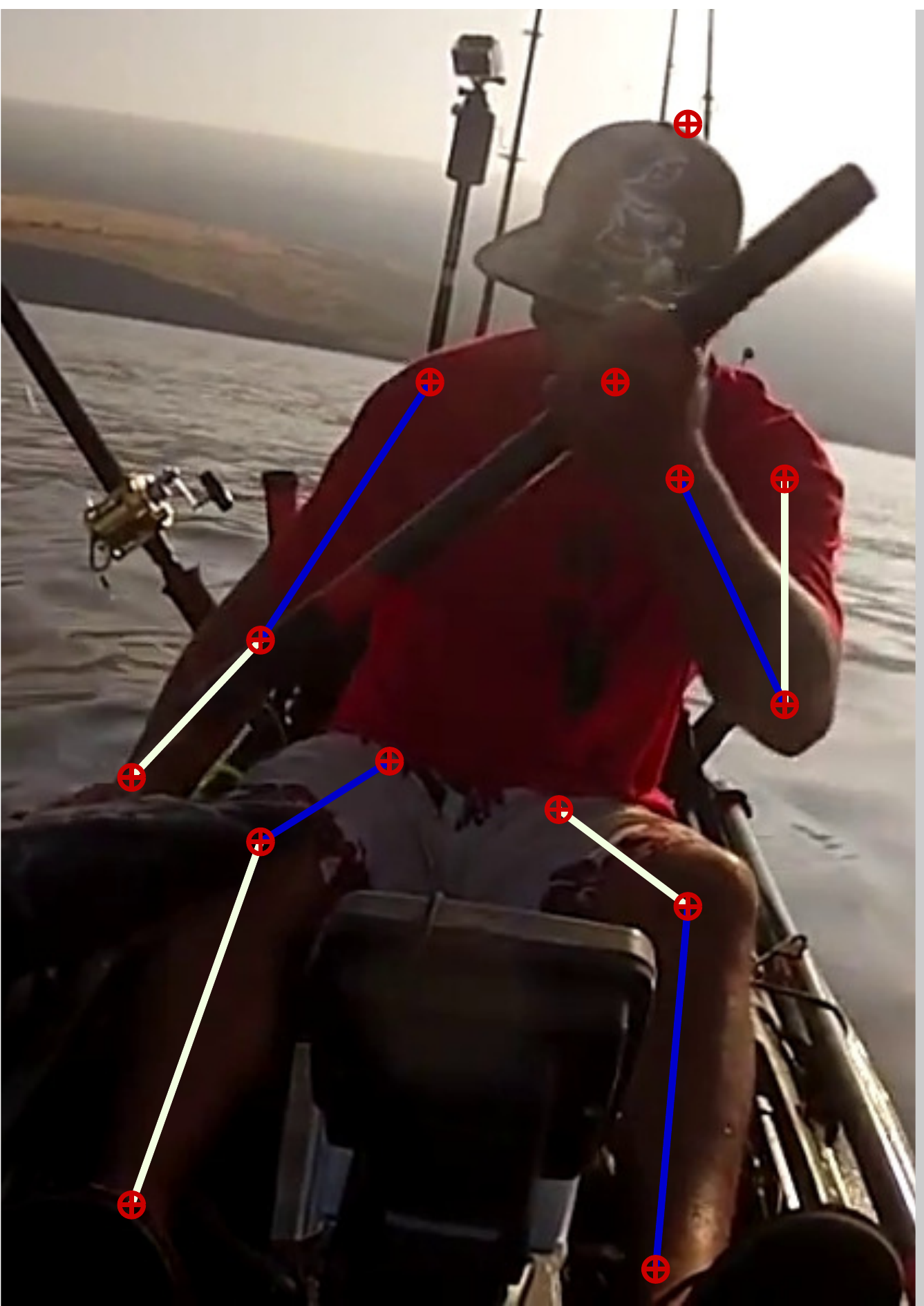}}
  \adjustbox{height=\myheightA}
      {\includegraphics[width=\textwidth]{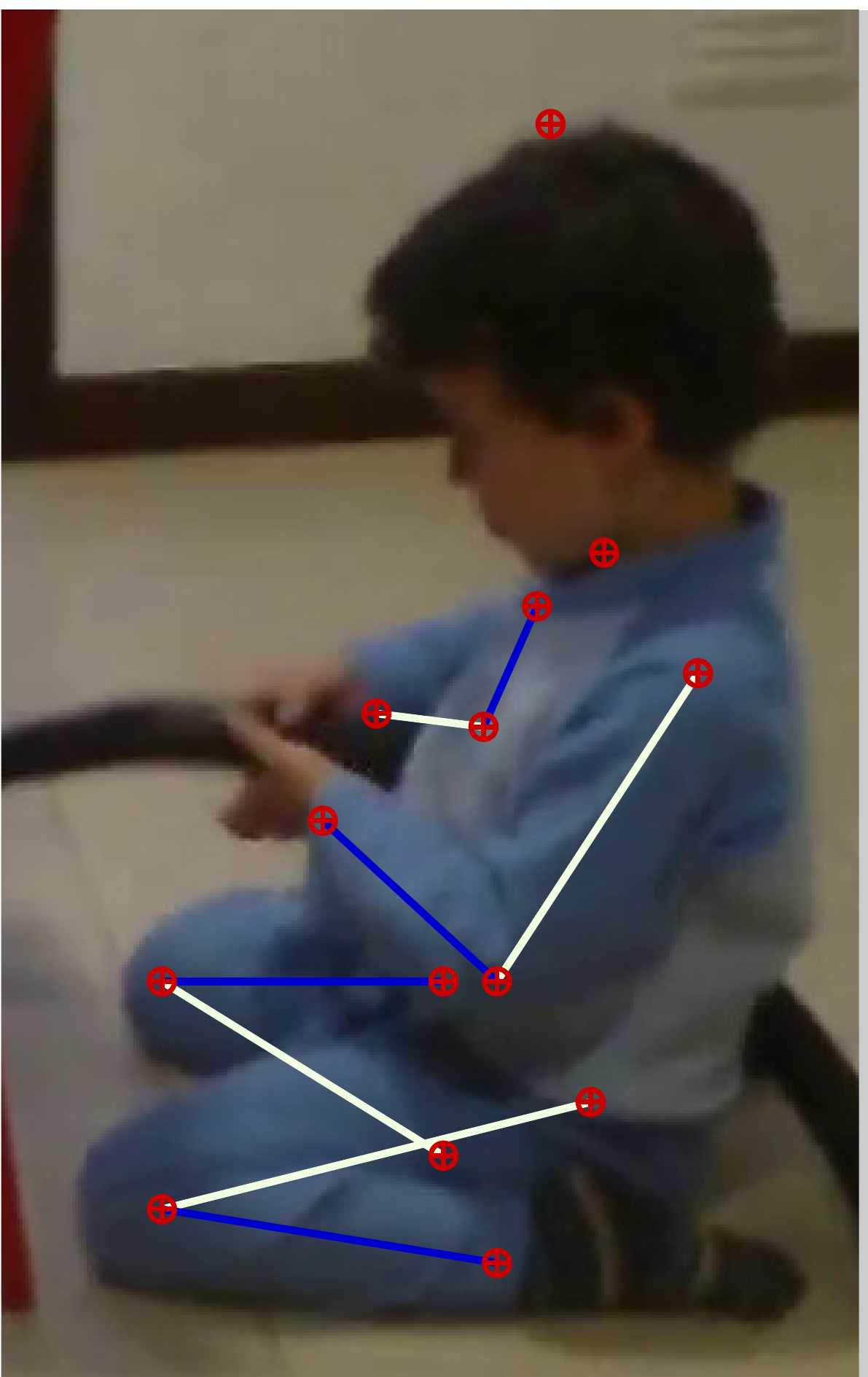}}
  \adjustbox{height=\myheightA}
      {\includegraphics[width=\textwidth]{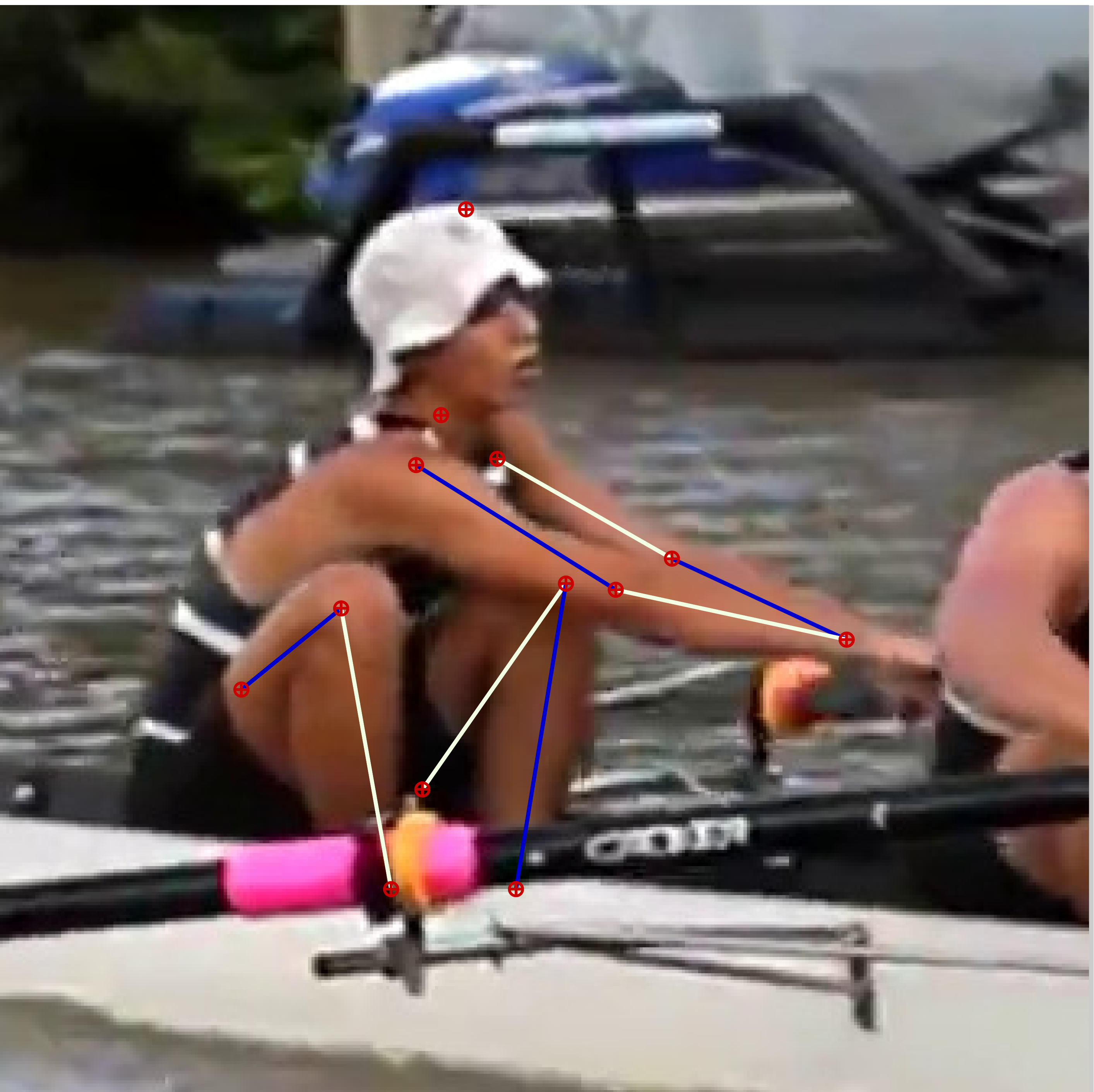}}
  \adjustbox{height=\myheightA}
      {\includegraphics[width=\textwidth]{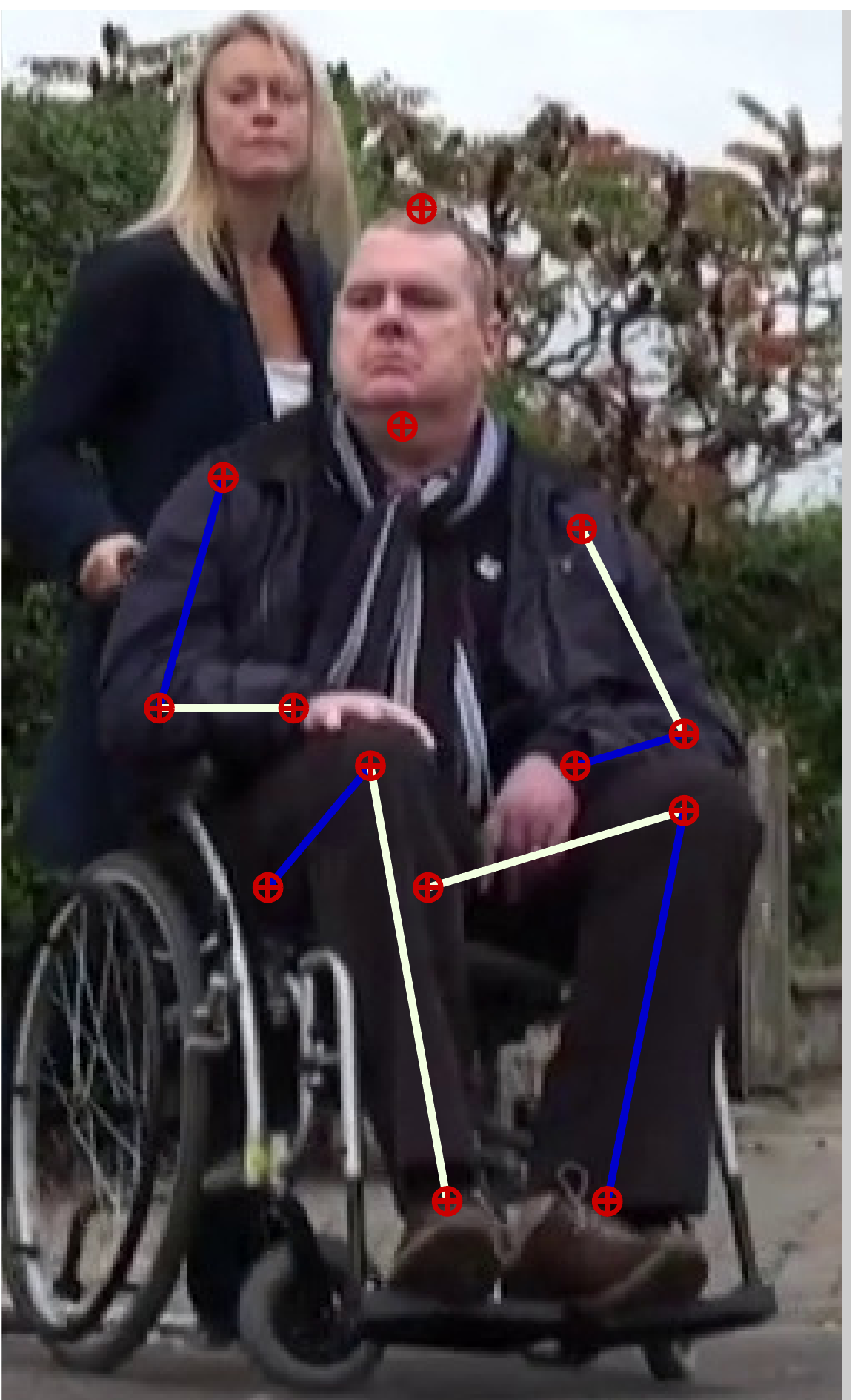}}
  \adjustbox{height=\myheightA}
      {\includegraphics[width=\textwidth]{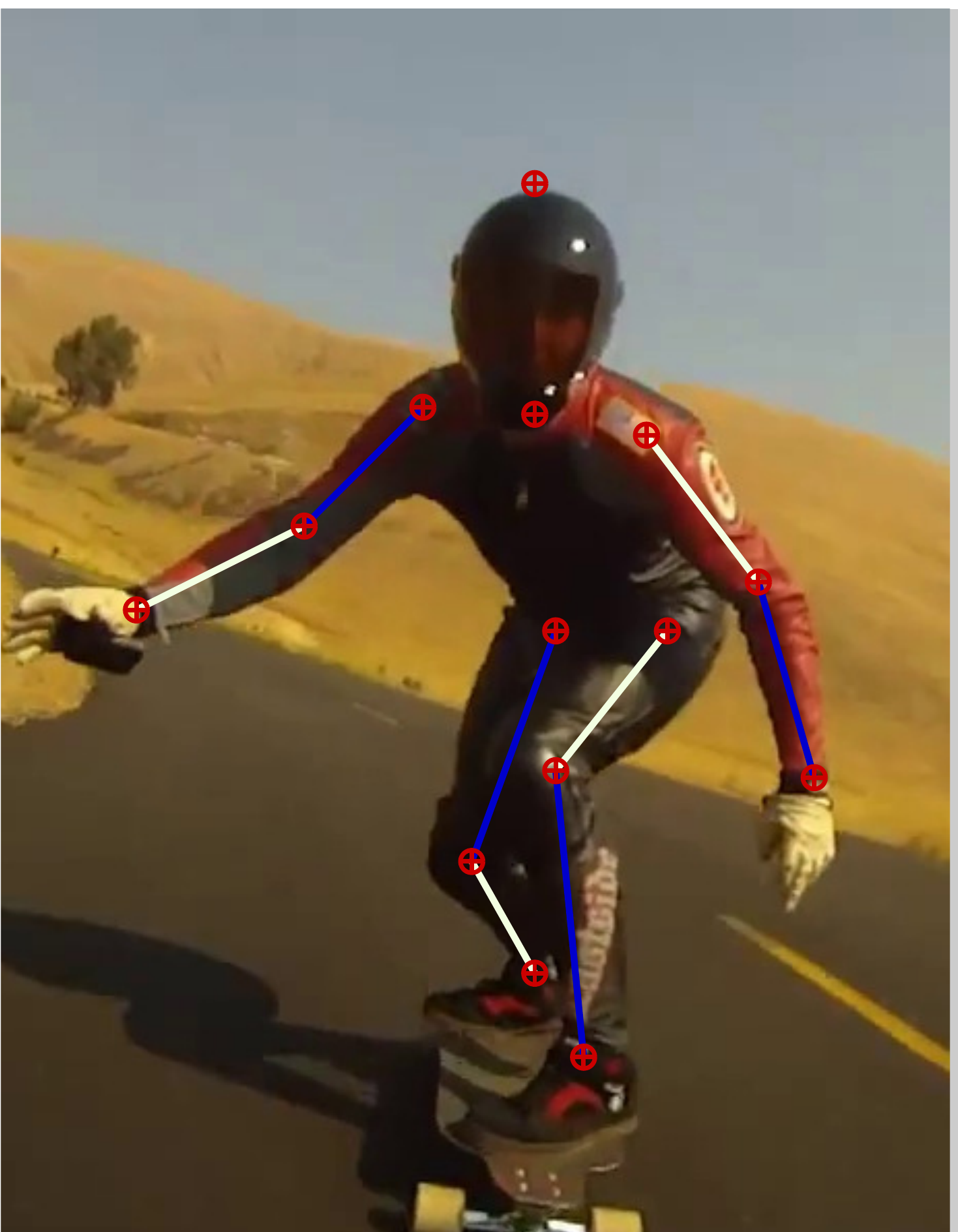}}

  \vspace{1mm}
  \centering
  \adjustbox{height=\myheightB}
      {\includegraphics[width=\textwidth]{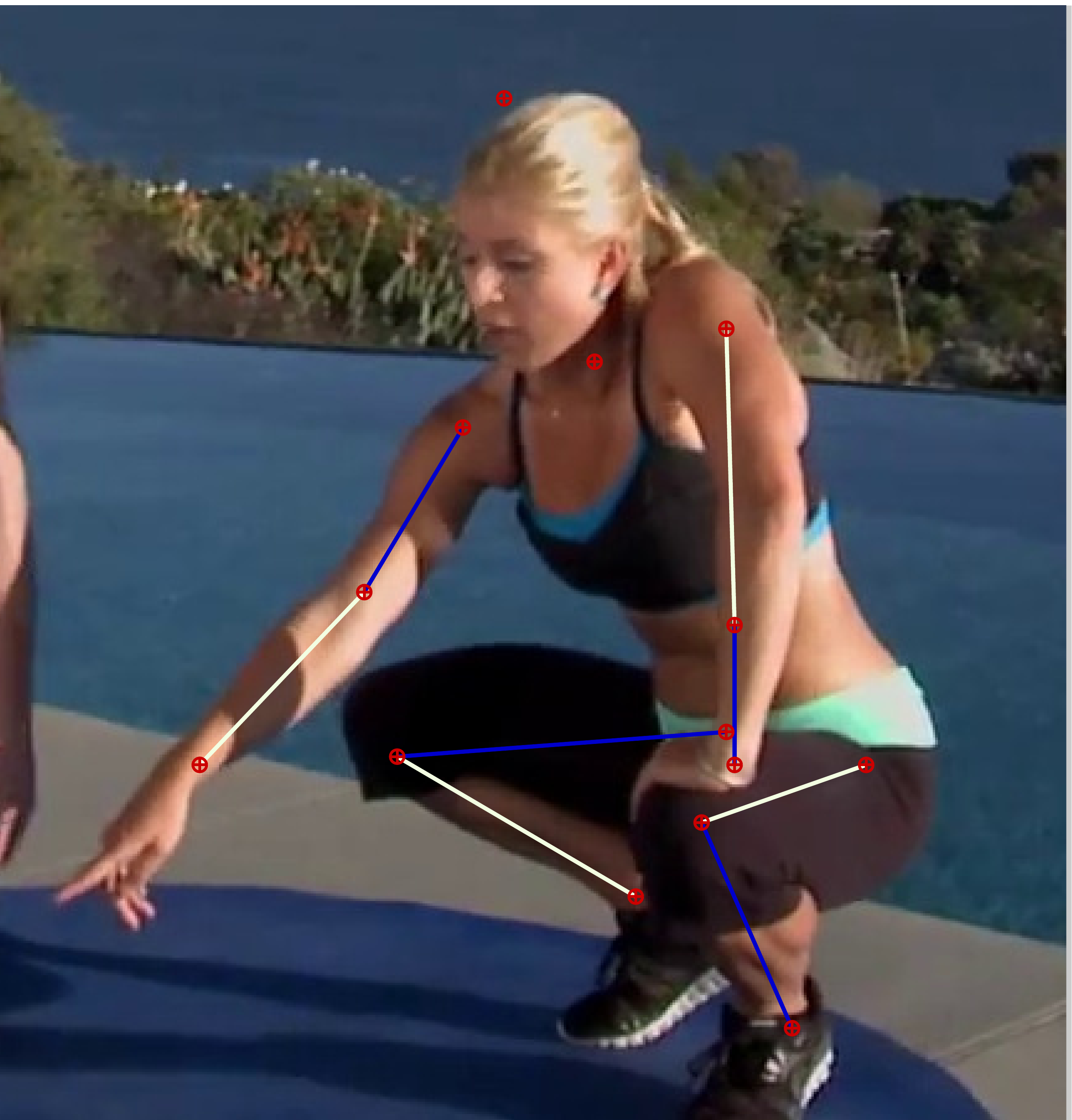}}
  \adjustbox{height=\myheightB}
      {\includegraphics[width=\textwidth]{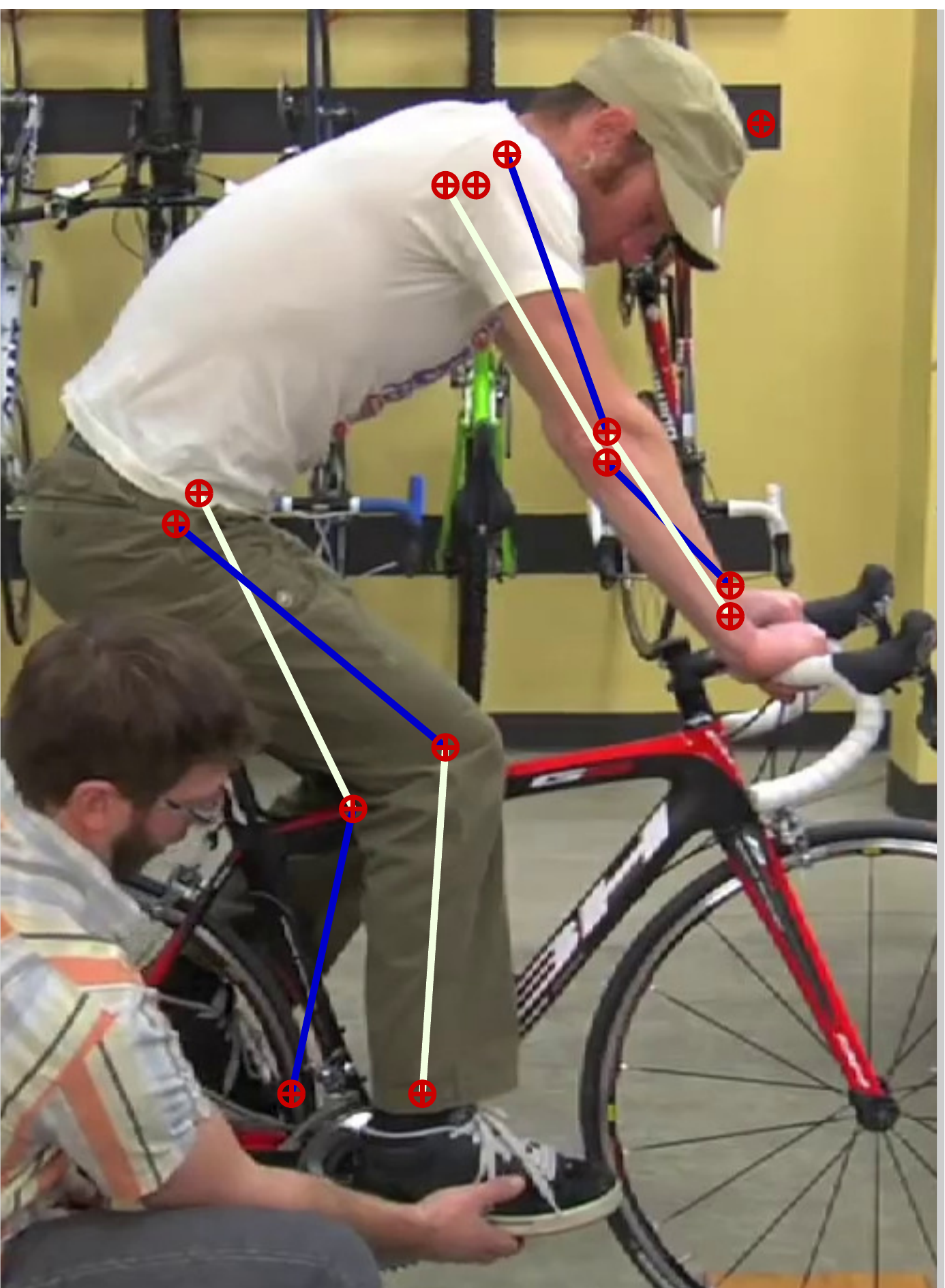}}
  \adjustbox{height=\myheightB}
      {\includegraphics[width=\textwidth]{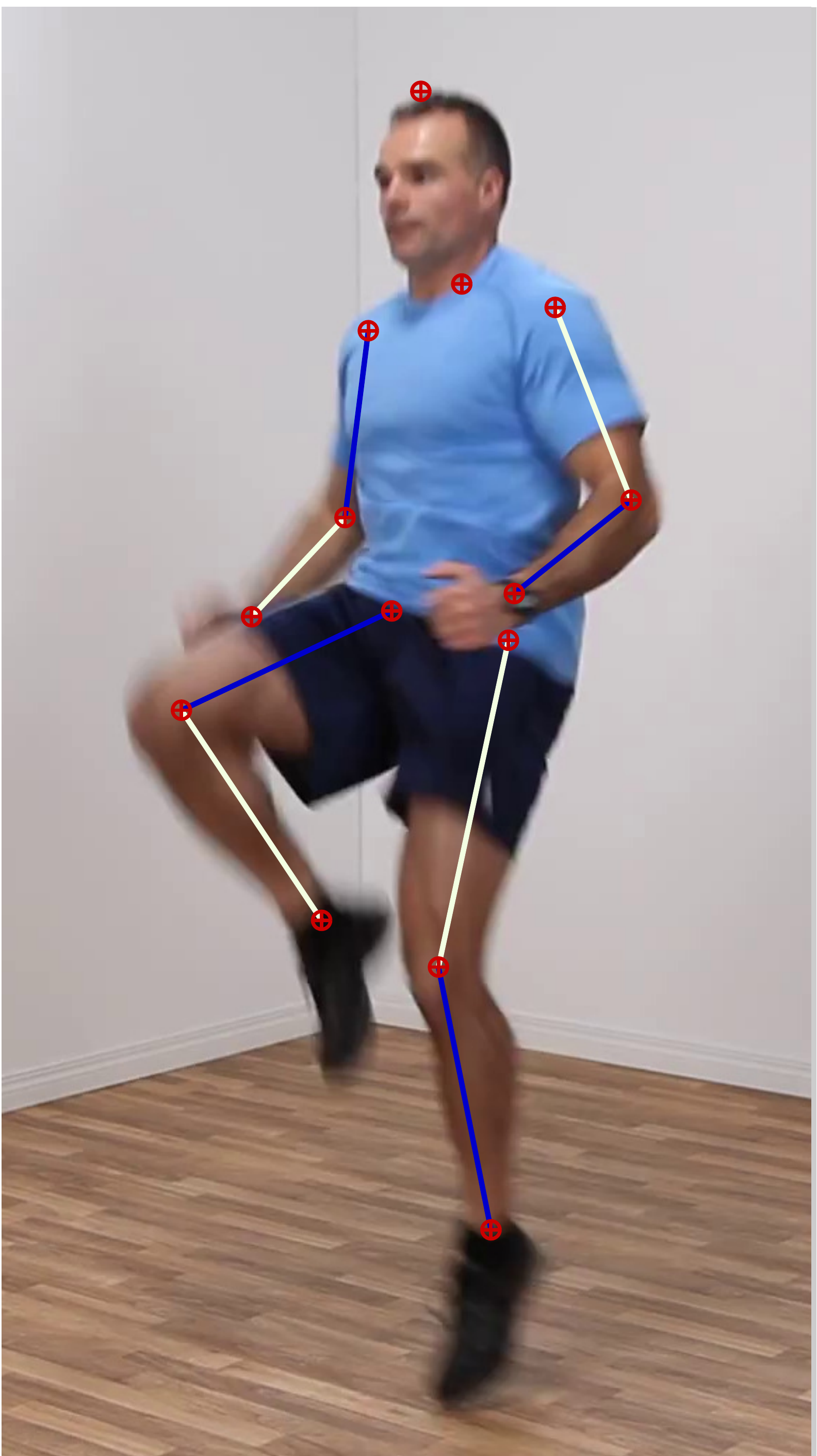}}
  \adjustbox{height=\myheightB}
      {\includegraphics[width=\textwidth]{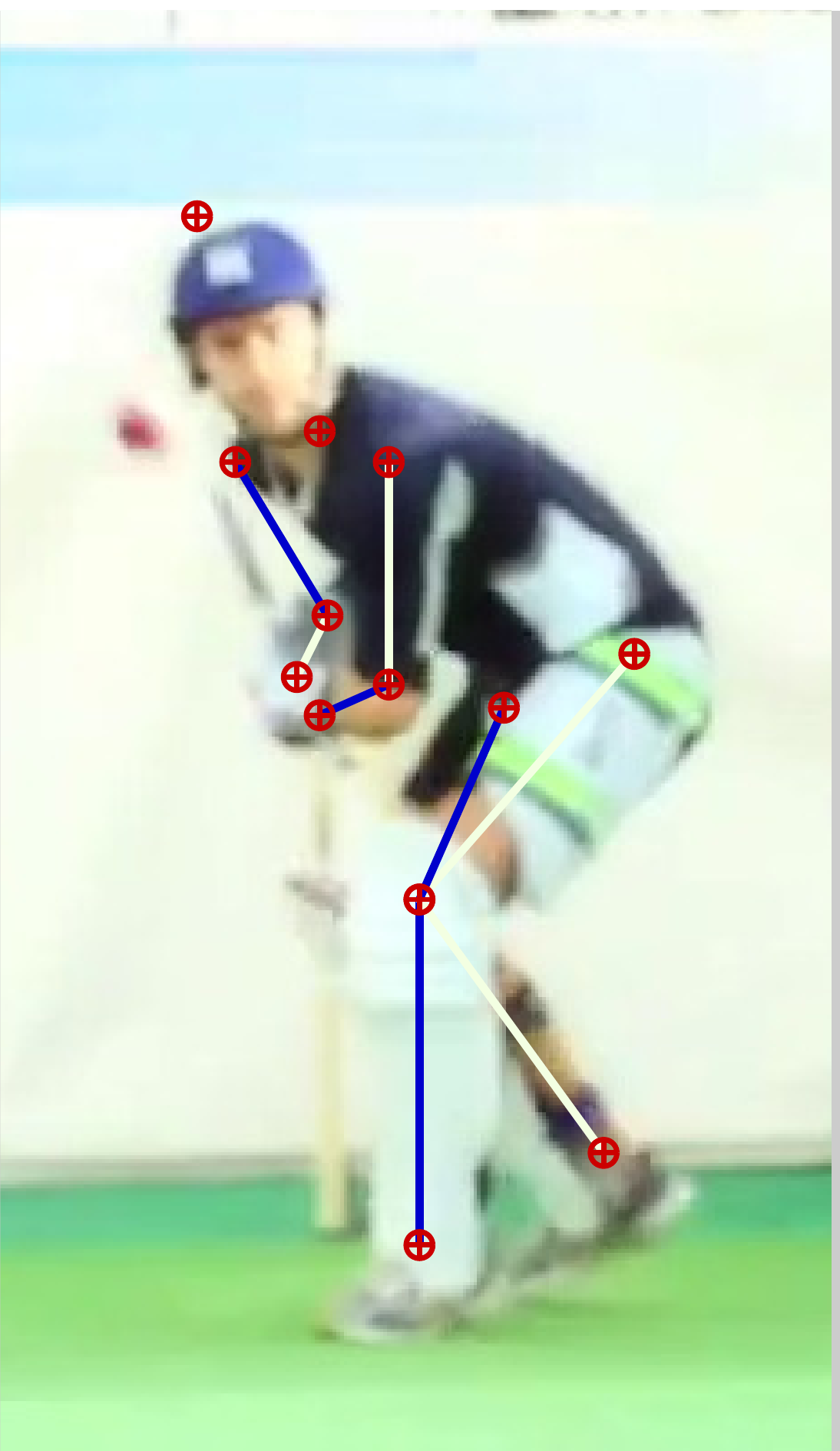}}
  \adjustbox{height=\myheightB}
      {\includegraphics[width=\textwidth]{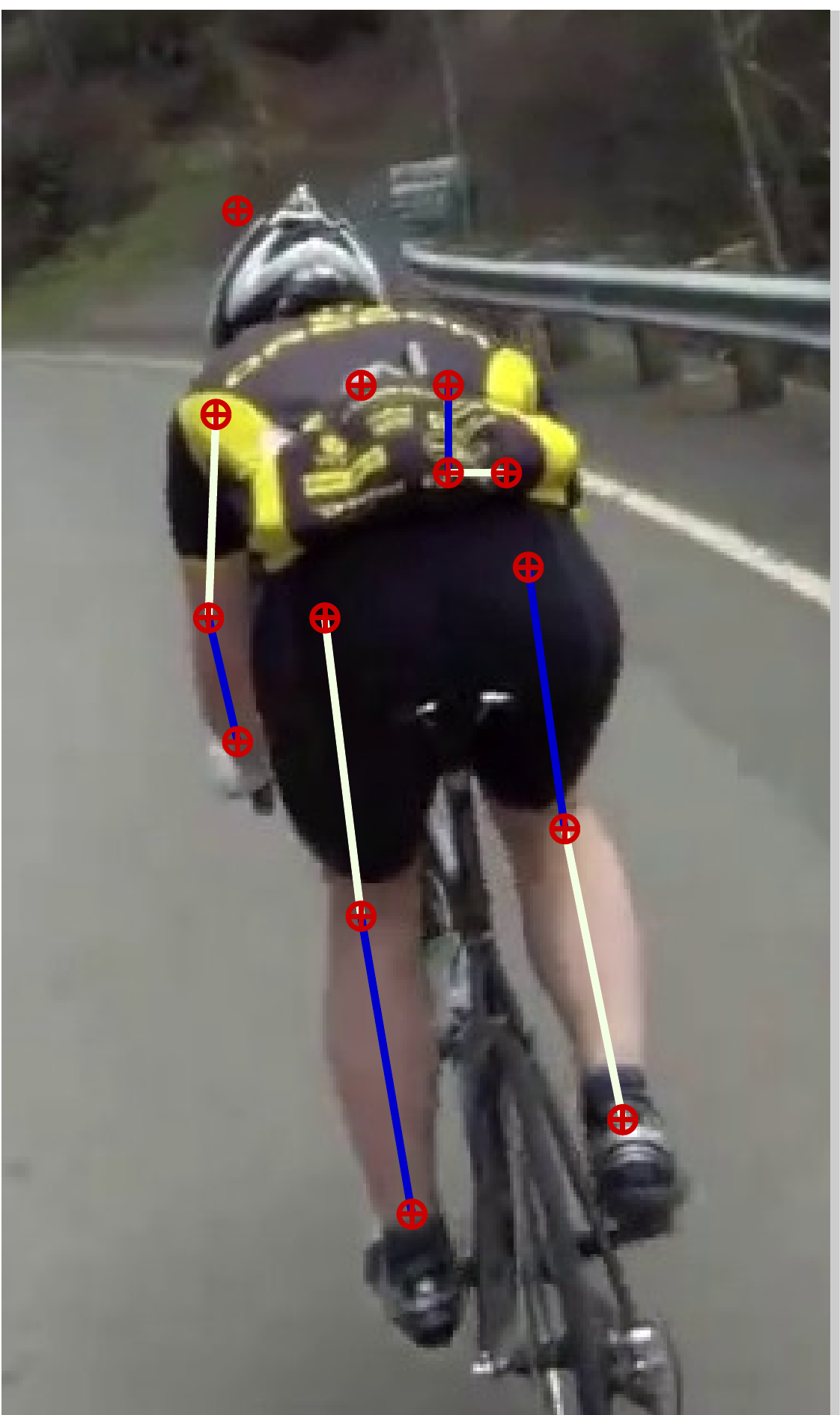}}
  \adjustbox{height=\myheightB}
      {\includegraphics[width=\textwidth]{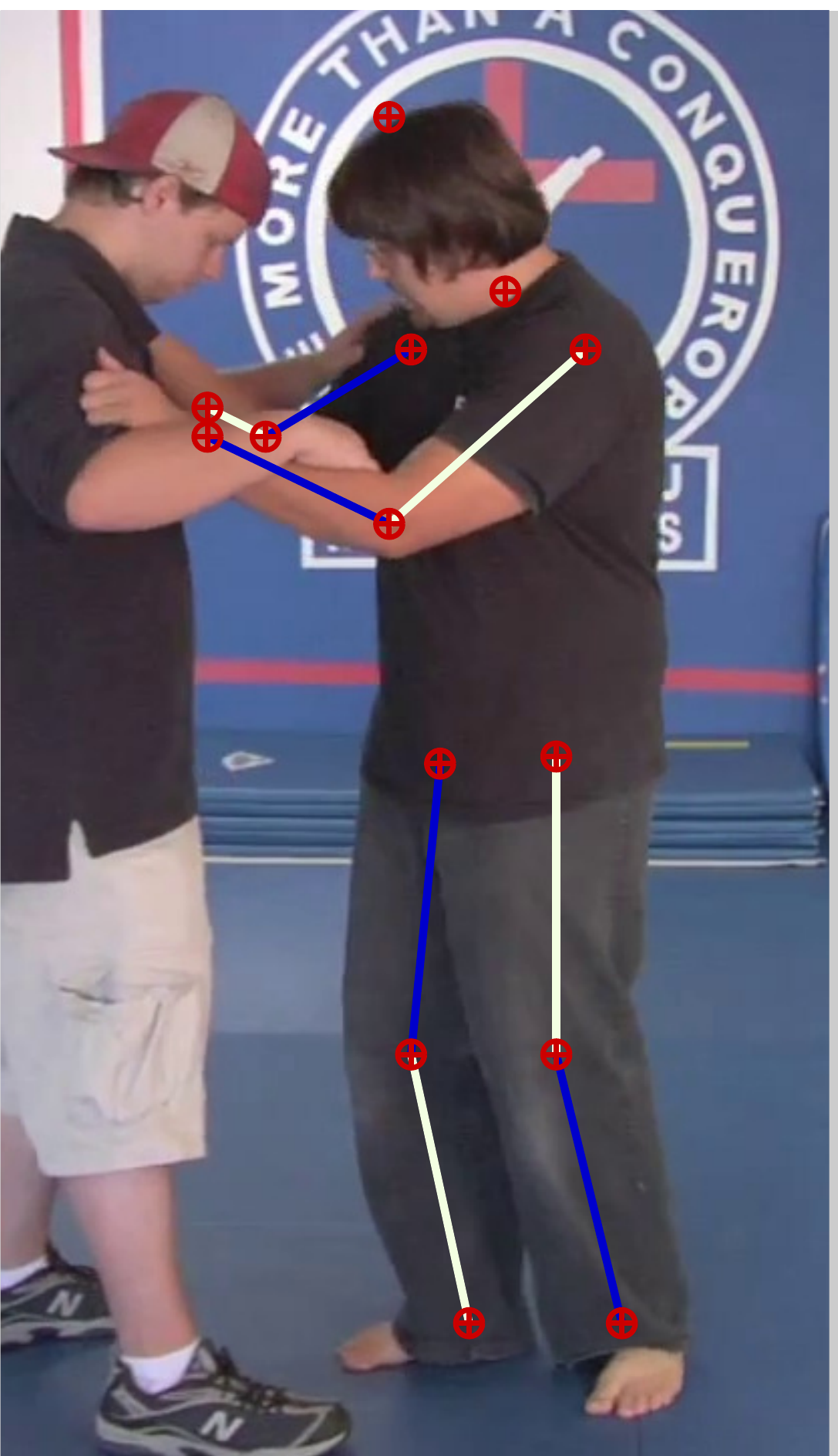}}
  \adjustbox{height=\myheightB}
      {\includegraphics[width=\textwidth]{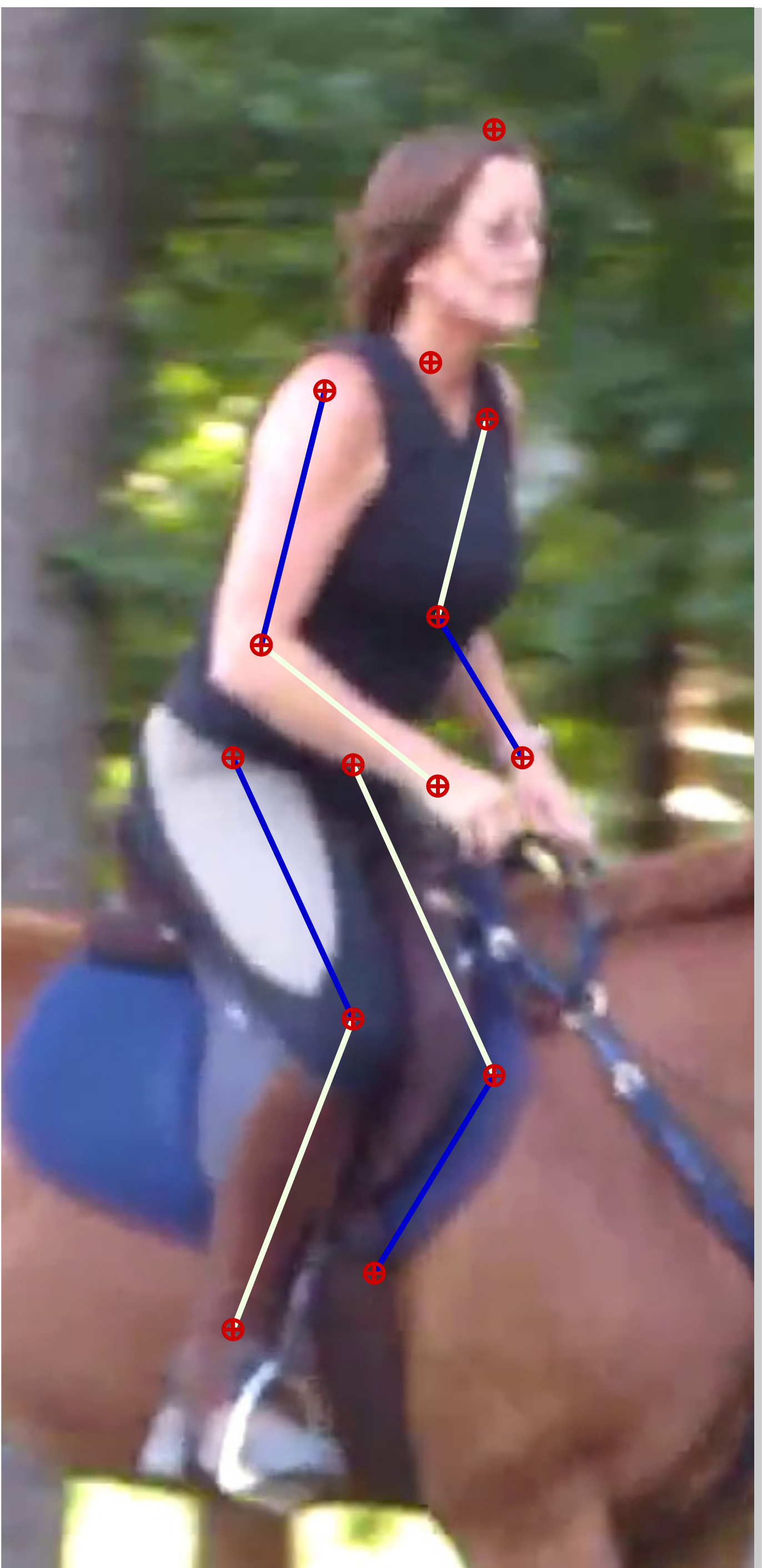}}

    \captionof{figure}{Our Model`s Predicted Joint Positions on the MPII-human-pose database test-set\cite{andriluka14cvpr}}
    \label{fig:pics}
\end{center}%
}]

\ifcvprfinal\thispagestyle{empty}\fi


\begin{abstract}
Recent state-of-the-art performance on human-body pose estimation has been achieved with Deep Convolutional Networks (ConvNets). Traditional ConvNet architectures include pooling and sub-sampling layers which reduce computational requirements, introduce invariance and prevent over-training. These benefits of pooling come at the cost of reduced localization accuracy. We introduce a novel architecture which includes an efficient `position refinement' model that is trained to estimate the joint offset location within a small region of the image. This refinement model is jointly trained in cascade with a state-of-the-art ConvNet model \cite{tompsonnips2014} to achieve improved accuracy in human joint location estimation. We show that the variance of our detector approaches the variance of human annotations on the FLIC~\cite{sapp13cvpr} dataset and outperforms all existing approaches on the MPII-human-pose dataset~\cite{andriluka14cvpr}.
\end{abstract}

\section{Introduction}

State-of-the-art performance on the task of human-body part localization has made significant progress in recent years. This has been in part due to the success of Deep-Learning architectures - specifically Convolutional Networks (ConvNets)~\cite{tompsonnips2014, arjunaccv2014, deeppose, chennips2014} - but also due to the availability of ever larger and more comprehensive datasets~\cite{andriluka14cvpr, Johnson10, sapp13cvpr} (our model's predictions for difficult examples from \cite{andriluka14cvpr} are shown in Figure~\ref{fig:pics}). 

A common characteristic of all ConvNet architectures used for human body pose detection to date is that they make use of internal strided-pooling layers. These layers reduce the spatial resolution by computing a summary statistic over a local spatial region (typically a max operation in the case of the commonly used Max-Pooling layer). The main motivation behind the use of these layers is to promote invariance to local input transformations (particularly translations) since their outputs are invariant to spatial location within the pooling region. This is particularly important for image classification where local image transformations obfuscate object identity. Therefore pooling plays a vital role in preventing over-training while reducing computational complexity for classification tasks.

The spatial invariance achieved by pooling layers comes at the price of limiting spatial localization accuracy. As such, by adjusting the amount of pooling in the network, for localization tasks a trade-off is made between generalization performance, model size and spatial accuracy.

In this paper we present a ConvNet architecture for efficient localization of human skeletal joints in monocular RGB images that achieves high spatial accuracy without significant computational overhead. This model allows us to use increased amounts of pooling for computational efficiency, while retaining high spatial precision.

We begin by presenting a ConvNet architecture to perform coarse body part localization. This network outputs a low resolution, per-pixel heat-map, describing the likelihood of a joint occurring in each spatial location. We use this architecture as a platform to discuss and empirically evaluate the role of Max-pooling layers in convolutional architectures for dimensionality reduction and improving invariance to noise and local image transformations. We then present a novel network architecture that reuses hidden-layer convolution features from the coarse heat-map regression model in order to improve localization accuracy. By jointly-training these models, we show that our model outperforms recent state-of-the-art on standard human body pose datasets~\cite{andriluka14cvpr, sapp13cvpr}.

\section{Related Work}

\begin{figure*}[ht]
\begin{center}
\includegraphics[width=0.8\textwidth]{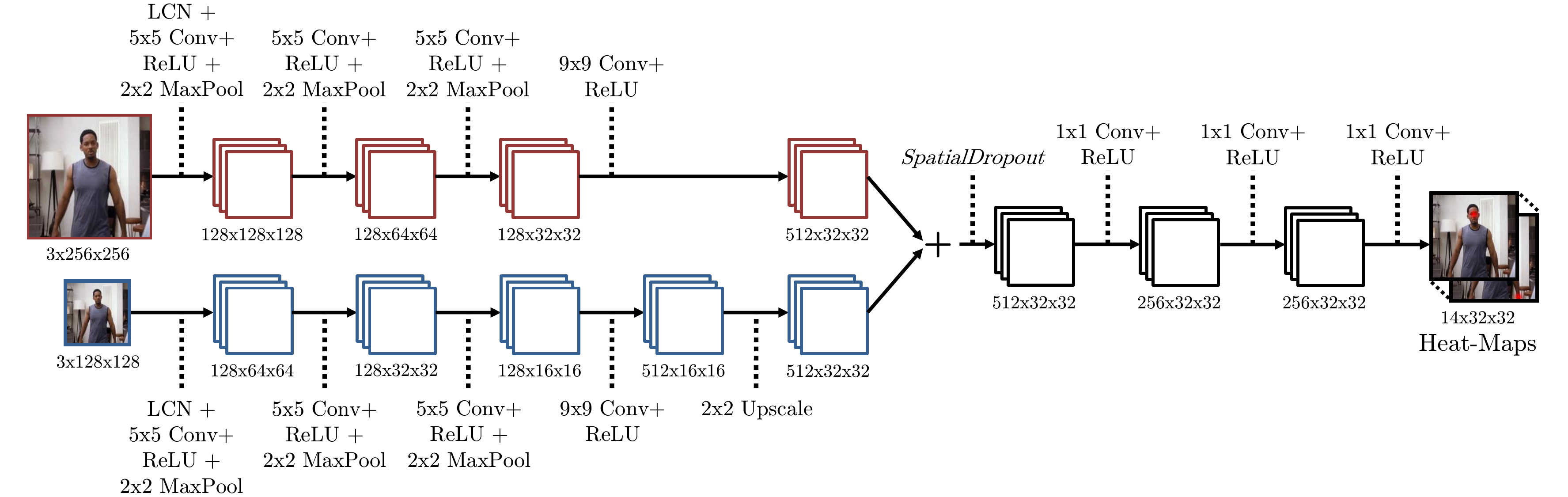}
\end{center}
   \caption{Multi-resolution Sliding Window Detector With Overlapping Contexts (model used on FLIC dataset)}
\label{fig:model}
\end{figure*}

Following the seminal work of Felzenszwalb et al.~\cite{felzenszwalb2008discriminatively} on `Deformable Part Models' (DPM) for human-body-pose estimation, many algorithms have been proposed to improve on the DPM architecture~\cite{andriluka2009pictorial, Eichner:2009:BAM, yang11cvpr,dantone13cvpr}. Yang and Ramanan~\cite{yang11cvpr} propose a mixture of templates modeled using SVMs. Johnson and Everingham~\cite{johnson11cvpr} propose more discriminative templates by using a cascade of body-part detectors. 

Recently high-order DPM-based body-part dependency models have been proposed~\cite{pishchulin13cvpr, pishchulin13iccv, Gkioxari2013, sapp13cvpr}. Pishchulin~\cite{pishchulin13cvpr, pishchulin13iccv} use \emph{Poselet} priors and a DPM model~\cite{PoseletsICCV09} to capture spatial relationships of body-parts. In a similar work, Gkioxari et al.~\cite{Gkioxari2013} propose the \emph{Armlets} approach which uses a semi-global classifier of part configurations.  Their approach exhibits good performance on real-world data, however it is demonstrated only on arms. Sapp and Taskar~\cite{sapp13cvpr} propose a multi-modal model including both holistic and local cues for coarse mode selection and pose estimation. A common characteristic to all these approaches is that they use hand-crafted features (edges, contours, HoG features and color histograms), which have been shown to have poor generalization performance and discriminative power in comparison to learned features (as in this work).

Today, the best performing algorithms for many vision tasks are based on convolutional networks (ConvNets). The current state-of-the-art methods for the task of human-pose estimation \emph{in-the-wild} are also built using ConvNets~\cite{deeppose, jainiclr2014, tompsonnips2014, arjunaccv2014, chennips2014}. The model of Toshev et al.~\cite{deeppose} significantly output-performed state-of-art methods on the challenging `FLIC' \cite{sapp13cvpr} dataset and was competitive on the `LSP' \cite{Johnson10} dataset. In contrast to our work, they formulate the problem as a direct (continuous) regression to joint location rather than a discrete heat-map output. However, their method performs poorly in the high-precision region and we believe that this is because the mapping from input RGB image to XY location adds unnecessary learning complexity which weakens generalization.

For example, direct regression does not deal gracefully with multi-modal outputs (where a valid joint is present in two spatial locations). Since the network is forced to produce a single output for a given regression input, the network does not have enough degrees of freedom in the output representation to afford small errors which we believe leads to over-training (since small outliers - due to for instance the presence of a valid body part - will contribute to a large error in XY).

Chen et al.~\cite{chennips2014} use a ConvNet to learn a low-dimensional representation of the input image and use an image dependent spatial model and show improvement over~\cite{deeppose}. Tompson et al.~\cite{tompsonnips2014} uses a multi-resolution ConvNet architecture to perform heat-map likelihood regression which they train jointly with a graphical model network to further promote joint consistency. In similar work, Jain et al.~\cite{arjunaccv2014} also uses a multi-resolution ConvNet architecture, but they add motion features to the network input to further improve accuracy. Our Heat-Map regression model is largely inspired by both these works with improvements for better localization accuracy. The contributions of this work can be seen as an extension of the architecture of \cite{tompsonnips2014}, where we attempt to overcome the limitations of pooling to improve the precision of the spatial locality.

In an unrelated application, Eigen et al.~\cite{eigen2014} predict depth by using a cascade of coarse to fine ConvNet models. In their work the coarse model is pre-trained and the model parameters are fixed when training the fine model. By contrast, in this work we suggest a novel shared-feature architecture which enables joint training of both models to improve generalization performance and which samples a subset of the feature inputs to improve runtime performance.

\section{Coarse Heat-Map Regression Model}

Inspired by the work of Tompson et al.~\cite{tompsonnips2014}, we use a multi-resolution ConvNet architecture (Figure~\ref{fig:model}) to implement a sliding window detector with overlapping contexts to produce a coarse heat-map output. Since our work is an extension of their model, we will only present a very brief overview of the architecture and explain our extensions to their model.

\subsection{Model Architecture}
\label{sec:heatmapmodel}

The coarse heat-map regression model takes as input an RGB Gaussian pyramid of 3 levels (in Figure~\ref{fig:model} only 2 levels are shown for brevity) and outputs a heat-map for each joint describing the per-pixel likelihood for that joint occurring in each output spatial location. We use an input resolution of 320x240 and 256x256 pixels for the FLIC~\cite{sapp13cvpr} and MPII~\cite{andriluka14cvpr} datasets respectively. The first layer of the network is a local-contrast-normalization (LCN) layer with the same filter kernel in each of the three resolution banks. 


Each LCN image is then input to a 7 stage multi-resolution convolutional network (11 stages for the MPII dataset model). Due to the presence of pooling the heat-map output is at a lower resolution than the input image. It should be noted that the last 4 stages (or 3 stages for the MPII dataset model) effectively simulate a fully-connected network for a target input patch size (which is typically a much smaller context than the input image). We refer interested readers to~\cite{tompsonnips2014} for more details.

\subsection{\textbf{\textit{SpatialDropout}}}

We improve the model of ~\cite{tompsonnips2014} by adding an additional dropout layer before the first 1x1 convolution layer in Figure~\ref{fig:model}. The role of dropout is to improve generalization performance by preventing activations from becoming strongly correlated~\cite{hinton2012improving}, which in turn leads to over-training. In the standard dropout implementation, network activations are ``dropped-out'' (by zeroing the activation for that neuron) during training with independent probability $p_{\text{drop}}$. At test time all activations are used, but a gain of $1-p_{\text{drop}}$ is multiplied to the neuron activations to account for the increase in expected bias.

In initial experiments, we found that applying standard dropout (where each convolution feature map activation is ``dropped-out'' independently) before the $1\times1$ convolution layer generally increased training time but did not prevent over-training. Since our network is fully convolutional and natural images exhibit strong spatial correlation, the feature map activations are also strongly correlated, and in this setting standard dropout fails. 

\begin{figure}[th]
\begin{center}
\includegraphics[width=0.9\columnwidth]{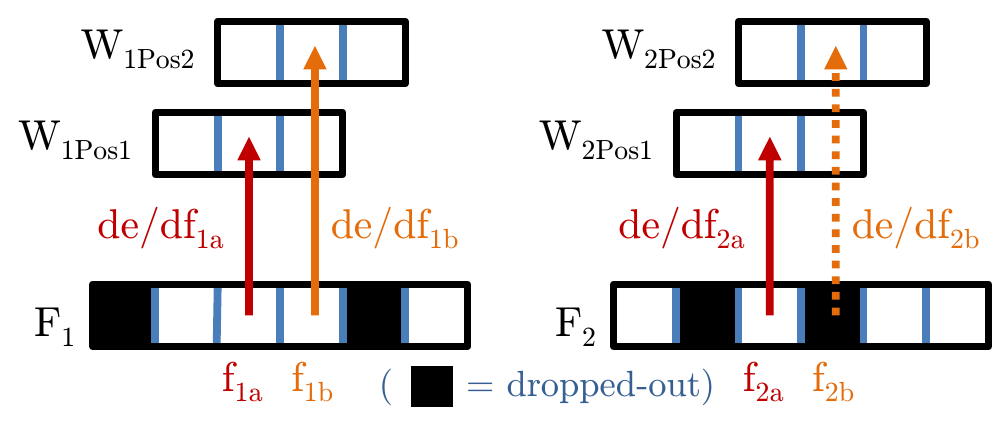}
\end{center}
   \caption{Standard Dropout after a 1D convolution layer}
\label{fig:dropout}
\end{figure}

Standard dropout at the output of a 1D convolution is illustrated in Figure~\ref{fig:dropout}. The top two rows of pixels represent the convolution kernels for feature maps 1 and 2, and the bottom row represents the output features of the previous layer. During back-propagation, the center pixel of the $W_2$ kernel receives gradient contributions from both $f_{2a}$ and $f_{2b}$ as the convolution kernel $W_2$ is translated over the input feature $F_2$. In this example $f_{2b}$ was randomly dropped out (so the activation was set to zero) while $f_{2a}$ was not. Since $F_2$ and $F_1$ are the output of a convolution layer we expect $f_{2a}$ and $f_{2b}$ to be strongly correlated: i.e. $f_{2a}\approx f_{2b}$ and $\nicefrac{de}{df_{2a}}\approx \nicefrac{de}{df_{2b}}$ (where $e$ is the error function to minimize). While the gradient contribution from $f_{2b}$ is zero, the strongly correlated $f_{2a}$ gradient remains. In essence, the effective learning rate is scaled by the dropout probability $p$, but independence is not enhanced.
\begin{figure*}[ht]
\begin{center}
\includegraphics[width=0.7\textwidth]{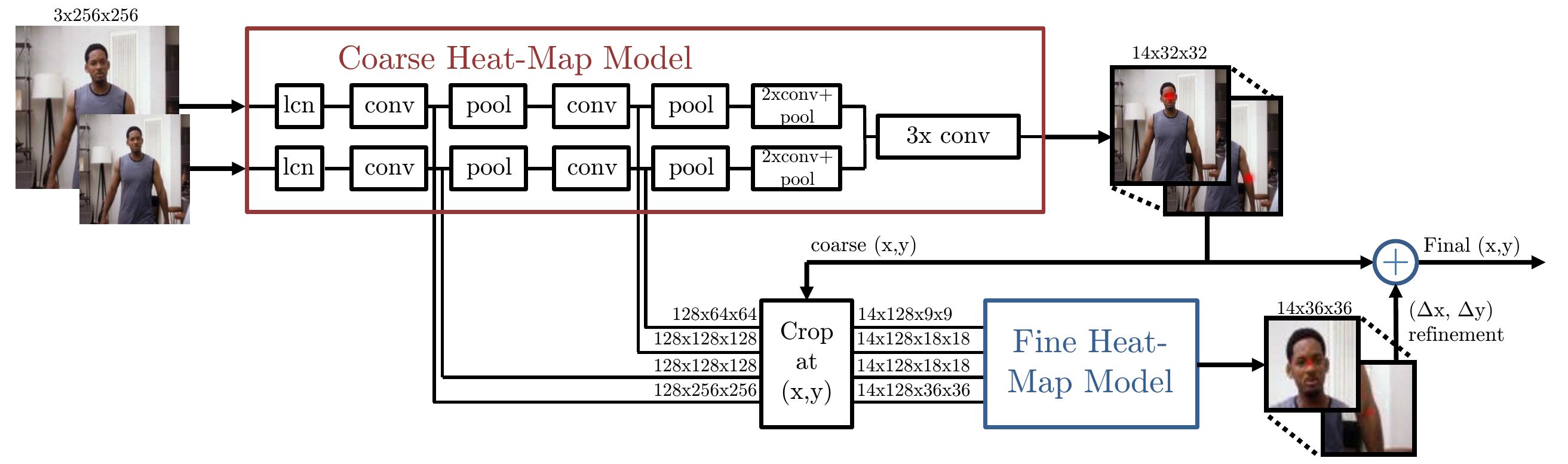}
\end{center}
   \caption{Overview of our Cascaded Architecture}
\label{fig:reg_overview}
\end{figure*}

\begin{figure}[th]
\begin{center}
\includegraphics[width=0.9\columnwidth]{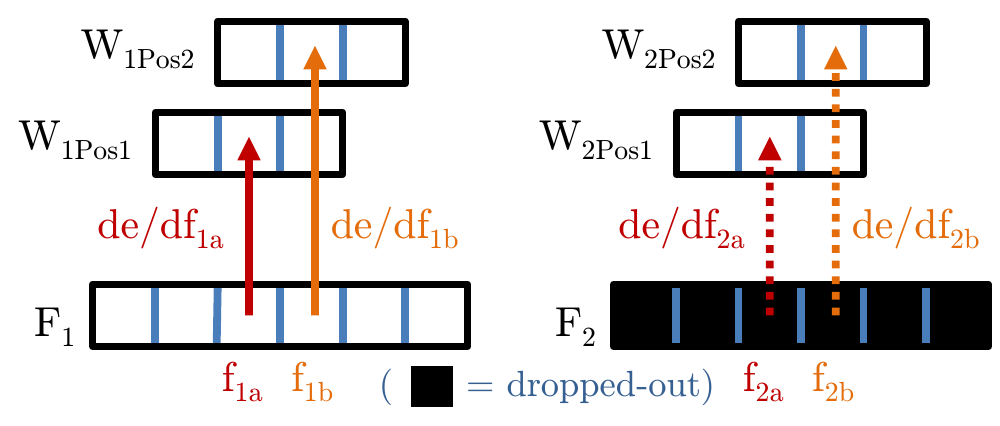}
\end{center}
   \caption{\textit{SpatialDropout} after a 1D convolution layer}
\label{fig:dropout_ours}
\end{figure}

Instead we formulate a new dropout method which we call \textit{SpatialDropout}. For a given convolution feature tensor of size $n_{\text{feats}}\times\text{height}\times\text{width}$, we perform only $n_{\text{feats}}$ dropout trials and extend the dropout value across the entire feature map. Therefore, adjacent pixels in the dropped-out feature map are either all $0$ (dropped-out) or all active as illustrated in Figure~\ref{fig:dropout_ours}. We have found this modified dropout implementation improves performance, especially on the FLIC dataset, where the training set size is small.

\subsection{Training and Data Augmentation}
 
We train the model in Figure~\ref{fig:model} by minimizing the Mean-Squared-Error (MSE) distance of our predicted heat-map to a target heat-map. The target is a 2D Gaussian of constant variance ($\sigma \approx 1.5$ pixels) centered at the ground-truth $(x,y)$ joint location. The objective function is:
\begin{equation}
E_1=\frac{1}{N}\sum_{j=1}^{N}{\sum_{xy}{\left\|H_{j}^\prime\left(x,y\right)-H_{j}\left(x,y\right)\right\|^2}}
\label{eq:objfuc}
\end{equation}

Where $H^\prime_j$ and $H_j$ are the predicted and ground truth heat-maps respectively for the $j$th joint.

During training, each input image is randomly rotated ($r\in[-20^{\circ},+20^{\circ}]$), scaled ($s\in[0.5,1.5]$) and flipped (with probability 0.5) in order to improve generalization performance on the validation-set.  Note that this follows the same training protocol as in ~\cite{tompsonnips2014}.

Many images contain multiple people while only a single person is annotated. To enable inference of the target person's annotations at test time, both the FLIC and MPII datasets include an approximate torso position. Since our sliding-window detector will detect all joint instances in a single frame indiscriminately, we incorporate this torso information by implementing the MRF-based spatial model of Tompson et al.~\cite{tompsonnips2014}, which formulates a tree-structured MRF over spatial locations with a random variable for each joint. The most likely joint locations are inferred (using message passing) given the noisy input distributions from the ConvNet. The ground-truth torso location is concatenated with the 14 predicted joints from the ConvNet output and these 15 joints locations are then input to the MRF. In this setup, the MRF inference step will learn to attenuate the joint activations from people for which the ground-truth torso is not anatomically viable, thus ``selecting" the correct person for labeling. Interested readers should refer to [20] for further details.

\section{Fine Heat-Map Regression Model}

In essence, the goal of this work is to recover the spatial accuracy lost due to pooling of the model in Section~\ref{sec:heatmapmodel} by using an additional ConvNet to refine the localization result of the coarse heat-map. However, unlike a standard cascade of models, as in the work of Toshev et al.~\cite{deeppose}, we reuse existing convolution features. This not only reduces the number of trainable parameters in the cascade, but also acts as a regularizer for the coarse heat-map model since the coarse and fine models are trained jointly.

\subsection{Model Architecture}
\label{sec:fine_model}

The full system architecture is shown in Figure~\ref{fig:reg_overview}. It consists of the heat-map-based parts model from Section~\ref{sec:heatmapmodel} for coarse localization, a module to sample and crop the convolution features at a specified $(x,y)$ location for each joint, as well as an additional convolutional model for fine tuning.

Joint inference from an input image is as follows: we forward-propagate (FPROP) through the coarse heat-map model then infer all joint $(x,y)$ locations from the maximal value in each joint's heat-map. We then use this coarse $(x,y)$ location to sample and crop the first 2 convolution layers (for all resolution banks) at each of the joint locations. We then FPROP these features through a fine heat-map model to produce a $(\Delta x,\Delta y)$ offset within the cropped sub-window. Finally, we add the position refinement to the coarse location to produce a final $(x,y)$ localization for each joint.

\begin{figure}[ht]
\begin{center}
\includegraphics[width=0.9\columnwidth]{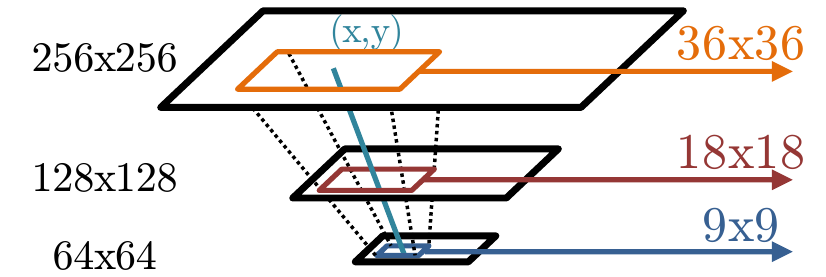}
\end{center}
   \caption{Crop module functionality for a single joint}
\label{fig:sampler}
\end{figure}

Figure~\ref{fig:sampler} shows the crop module functionality for a single joint. We simply crop out a window centered at the coarse joint $(x,y)$ location in each resolution feature map, however we do so by keeping the contextual size of the window constant by scaling the cropped area at each higher resolution level. Note that back-propagation (BPROP) through this module from output feature to input feature is trivial; output gradients from the cropped image are simply added to the output gradients of the convolution stages in the coarse heat-map model at the sampled pixel locations.

The fine heat-map model is a Siamese network~\cite{bromley1993signature} of 7 instances (14 for the MPII dataset), where the weights and biases of each module are shared (i.e. replicated across all instances and updated together during BPROP). Since the sample location for each joint is different, the convolution features do not share the same spatial context and so the convolutional sub-networks must be applied to each joint independently. However, we use parameter sharing amongst each of the 7 instances to substantially reduce the number of shared parameters and to prevent over-training. At the output of each of the 7 sub-networks we then perform a 1x1 Convolution, with no weight sharing to output a detailed-resolution heat-map for each joint. The purpose of this last layer is to perform the final detection for each joint.
\begin{figure}[th]
\begin{center}
\includegraphics[width=0.9\columnwidth]{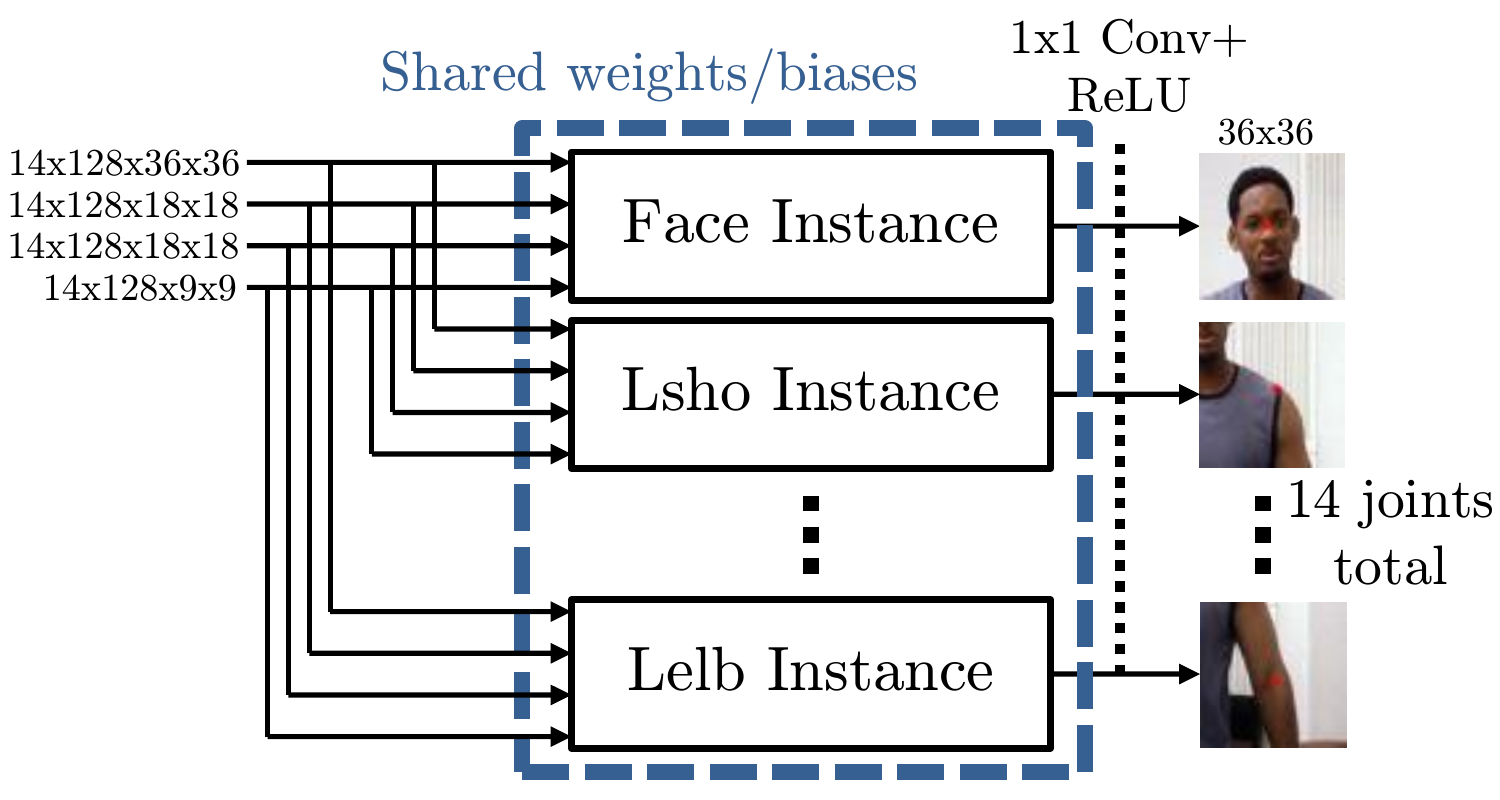}
\end{center}
   \caption{Fine heat-map model: 14 joint Siamese network}
\label{fig:reg_network_siamese}
\end{figure}

\begin{figure}[th]
\begin{center}
\includegraphics[width=1.0\columnwidth]{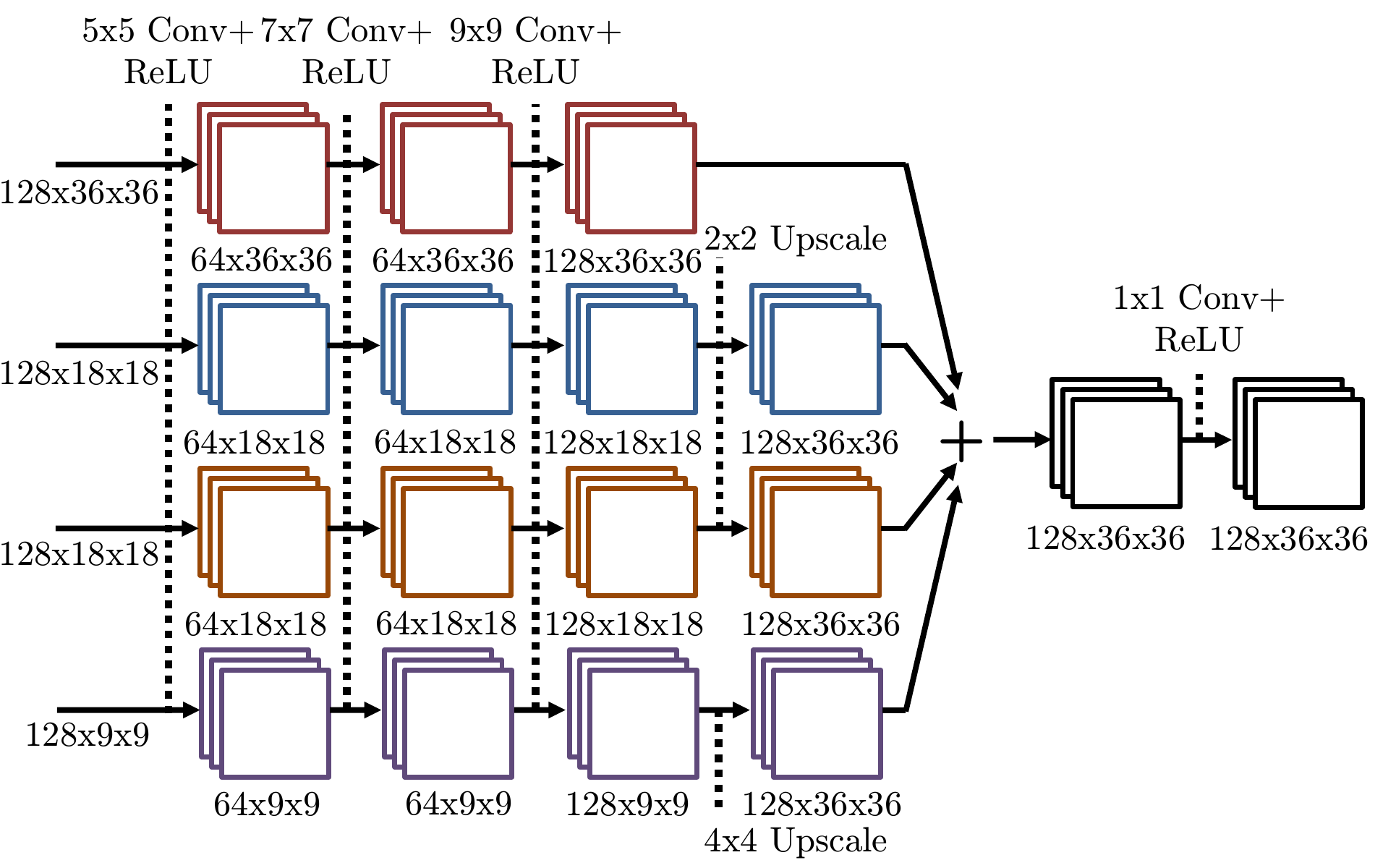}
\end{center}
   \caption{The fine heat-map network for a single joint}
\label{fig:reg_network}
\end{figure}

Note we are potentially performing redundant computations in the Siamese network. If two cropped sub-windows overlap and since the convolutional weights are shared, the same convolution maybe applied multiple times to the same spatial locations. However, we have found in practice this is rare. Joints are infrequently co-located, and the spatial context size is chosen such that there is little overlap between cropped sub-regions (note that the context of the cropped images shown in Figures~\ref{fig:reg_overview} and ~\ref{fig:reg_network} are exaggerated for clarity).

Each instance of the sub-network in Figure~\ref{fig:reg_network_siamese} is a ConvNet of 4 layers, as shown in Figure~\ref{fig:reg_network}. Since the input images are different resolutions and originate from varying depths in the coarse heat-map model, we treat the input features as separate resolution banks and apply a similar architecture strategy as used in Section~\ref{sec:heatmapmodel}. That is we apply the same size convolutions to each bank, upscale the lower-resolution features to bring them into canonical resolution, add the activations across feature maps then apply 1x1 convolutions to the output features.

It should be noted that this cascaded architecture can be extended further as is possible to have multiple cascade levels each with less and less pooling. However, in practice we have found that a single layer provides sufficient accuracy, and in particular within the level of label noise on the FLIC dataset (as we show in Section~\ref{sec:results}).

\subsection{Joint Training}

Before joint training, we first pre-train the coarse heat-map model of Section~\ref{sec:heatmapmodel} by minimizing Eq~\ref{eq:objfuc}. We then hold the parameters of the coarse model fixed and train the fine heat-map model of Section~\ref{sec:fine_model} by minimizing:

\begin{equation}
E_2=\frac{1}{N}\sum_{j=1}^{N}{\sum_{x,y}{\left\|G_{j}^\prime\left(x,y\right)-G_{j}\left(x,y\right)\right\|^2}}
\end{equation}

Where $G^\prime$ and $G$ are the set of predicted and ground truth heat-maps respectively for the fine heat-map model. Finally, we jointly train both models by minimizing $E_3=E_1 + \lambda E_2$. Where $\lambda$ is a constant used to trade-off the relative importance of both sub-tasks. We treat $\lambda$ as another network hyper-parameter and is chosen to optimize performance over our validation set (we use $\lambda=0.1$). Ideally, a more direct optimization function would attempt to measure the $argmax$ of both heat-maps and therefore directly minimize the final $(x,y)$ prediction. However, since the $argmax$ function is not differentiable we instead reformulate the problem as a regression to a set of target heat-maps and minimize the distance to those heat-maps.

\section{Results}
\label{sec:results}

Our ConvNet architecture was implemented within the Torch7~\cite{torch7} framework and evaluation is performed on the FLIC~\cite{sapp13cvpr} and MPII-Human-Pose~\cite{andriluka14cvpr} datasets. The FLIC dataset consists of 3,987 training examples and 1,016 test examples of still scenes from Hollywood movies annotated with upper-body joint labels. Since the poses are predominantly front-facing and upright, FLIC is considered to be less challenging than more recent datasets. However the small number of training examples makes the dataset a good indicator for generalization performance. On the other-hand the MPII dataset is very challenging and it includes a wide variety of full-body pose annotations within the 28,821 training and 11,701 test examples. For evaluation of our model on the FLIC dataset we use the standard PCK measure proposed by \cite{sapp13cvpr} and we use the PCKh measure of \cite{andriluka14cvpr} for evaluation on the MPII dataset.

\begin{figure}[th]
\begin{center}
\includegraphics[width=0.75\columnwidth]{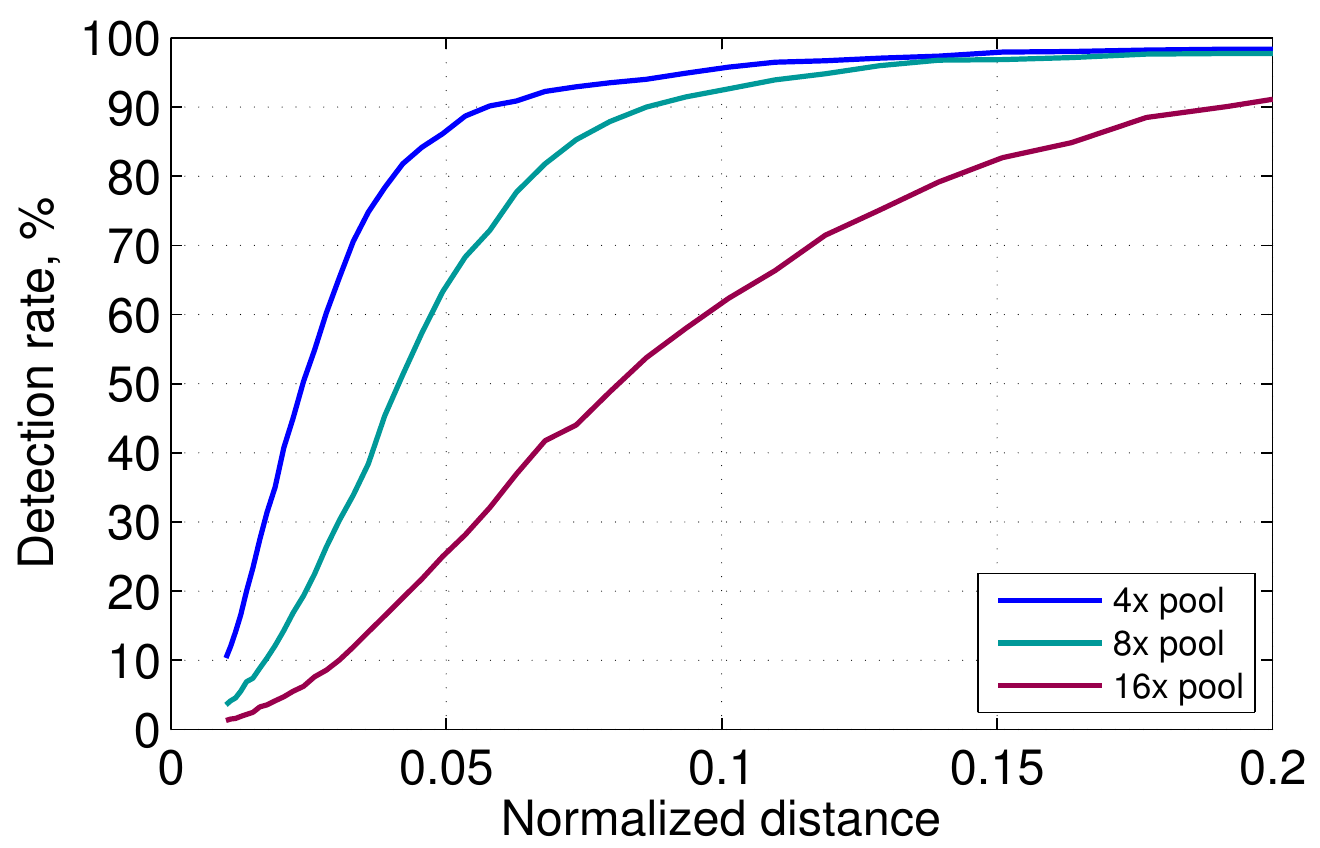}
\end{center}
   \caption{Pooling impact on FLIC test-set Average Joint Accuracy for the coarse heat-map model}
\label{fig:pooling}
\end{figure}

Figure~\ref{fig:pooling} shows the PCK test-set performance of our coarse heat-map model (Section~\ref{sec:heatmapmodel}) when various amounts of pooling are used within the network (keeping the number of convolution features constant). Figure~\ref{fig:pooling} results show quite clearly the expected effect of coarse quantization in $(x,y)$ and therefore the impact of pooling on spatial precision; when more pooling is used the performance of detections within small distance thresholds is reduced.

\begin{figure}[th]
  \centering
  \begin{subfigure}[b]{0.49\linewidth}
    \includegraphics[trim=250 100 60 65,clip,width=\textwidth]{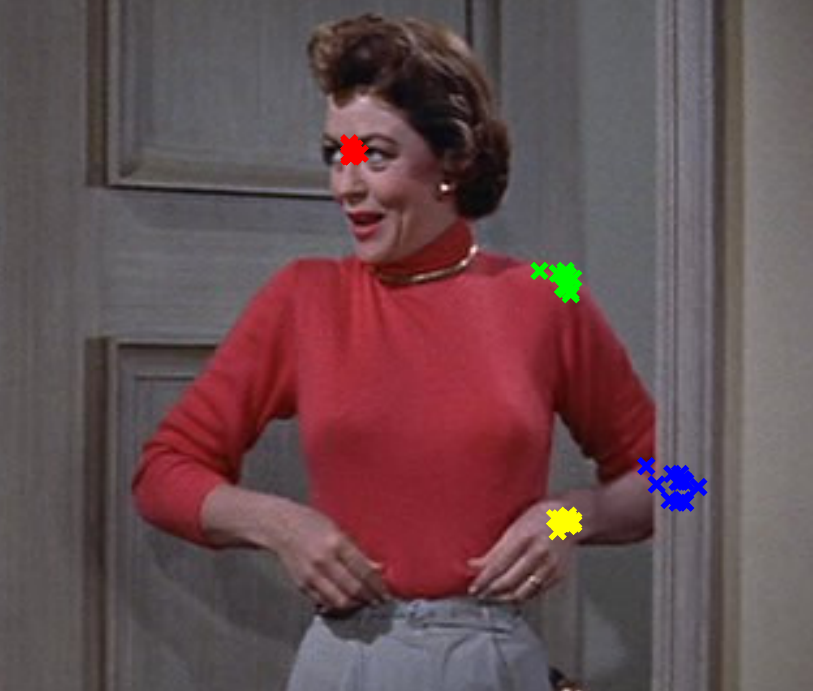}
  \end{subfigure}
  \begin{subfigure}[b]{0.49\linewidth}
    \includegraphics[trim=190 100 120 65,clip,width=\textwidth]{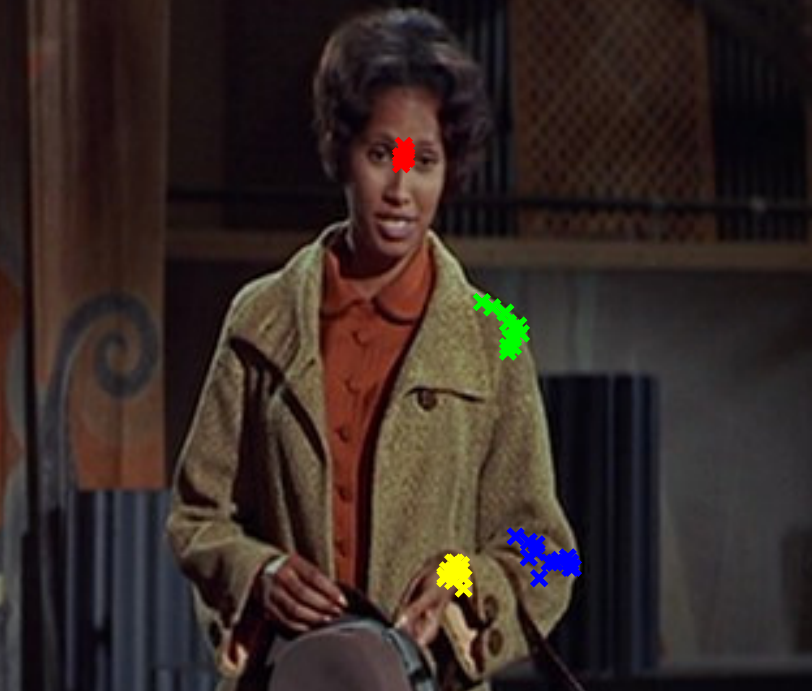}
  \end{subfigure}
  \caption{User generated joint annotations}
  \label{fig:jnt_labels}
\end{figure}

For joints where the ground-truth label is ambiguous and difficult for the human mechanical-turkers to label, we do not expect our cascaded network to do better than the expected variance in the user-generated labels. To measure this variance (and thus estimate the upper bound of performance) we performed the following informal experiment: we showed 13 users 10 random images from the FLIC training set with annotated ground-truth labels as a reference so that the users could familiarize themselves with the desired anatomical location of each joint. The users then annotated a consistent set of 10 random images from the FLIC test-set for the face, left-wrist, left-shoulder and left-elbow joints. Figure~\ref{fig:jnt_labels} shows the resultant joint annotations for 2 of the images.

To estimate joint annotation noise we calculate the standard deviation ($\sigma$) across user annotations in $x$ for each of the 10 images separately and then average the $\sigma$ across the 10 sample images to obtain an aggregate $\sigma$ for each joint. Since we down-sample the FLIC images by a factor of 2 for use with our model we divide the $\sigma$ values by the same down-sample ratio. The result is shown in Table~\ref{tab:noise}.
\begin{table}
\begin{center}
\begin{scriptsize} 
\setlength{\tabcolsep}{0.2pt}
\begin{tabular}{ l x{0.7cm} x{1.3cm} x{0.85cm} x{0.75cm} }
  \hline
  \noalign{\vskip 1mm}
  
                                    & Face & Shoulder & Elbow & Wrist \\
  \noalign{\vskip 1mm}
  \hline
  \noalign{\vskip 1mm}
                     Label Noise (10 images) & 0.65 & 2.46     & 2.14  & 1.57 \\
                     This work 4x (test-set)  & 1.09 & 2.43     & 2.59  & 2.82 \\
                     This work 8x (test-set)   & 1.46 & 2.72     & 2.49  & 3.41 \\
                     This work 16x (test-set)  & 1.45 & 2.78     & 3.78  & 4.16 \\                 
  \noalign{\vskip 1mm}
  \hline
\end{tabular}
\end{scriptsize}
\end{center}
\caption{$\sigma$ of $(x,y)$ pixel annotations on FLIC test-set images (at $360\times240$ resolution)}
\label{tab:noise}
\end{table}

\begin{figure}[th]
  \centering
  \begin{subfigure}[b]{0.49\linewidth}
        \includegraphics[width=\textwidth]{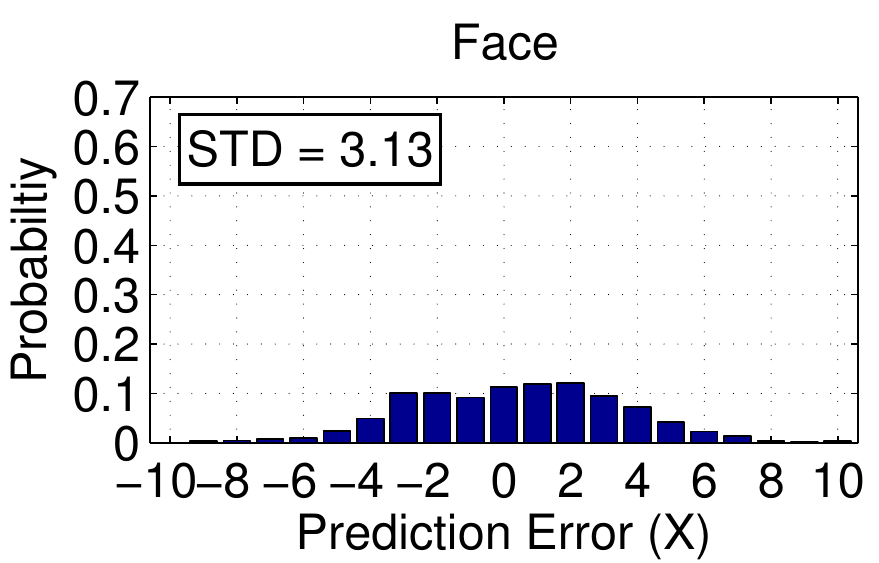}
  \end{subfigure}
  \begin{subfigure}[b]{0.49\linewidth}
        \includegraphics[width=\textwidth]{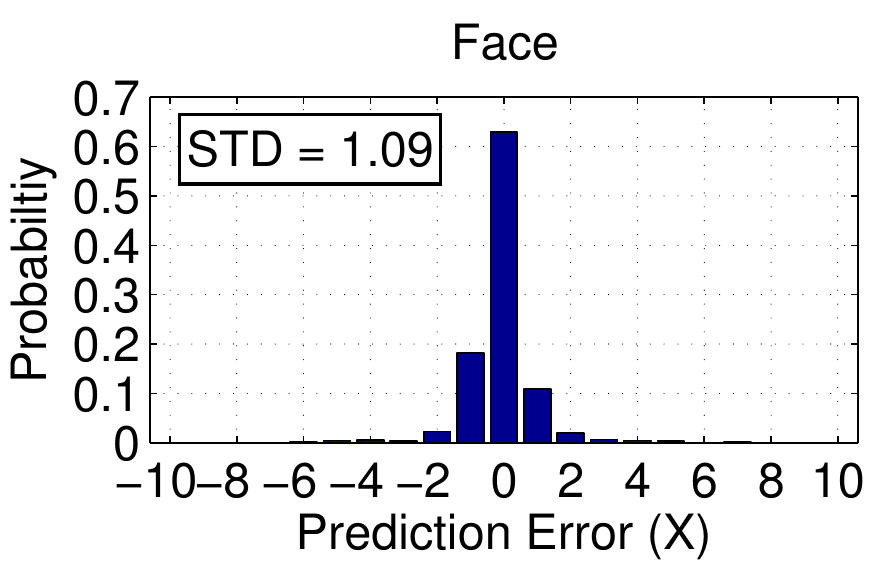}
  \end{subfigure}
  \begin{subfigure}[b]{0.49\linewidth}
        \includegraphics[width=\textwidth]{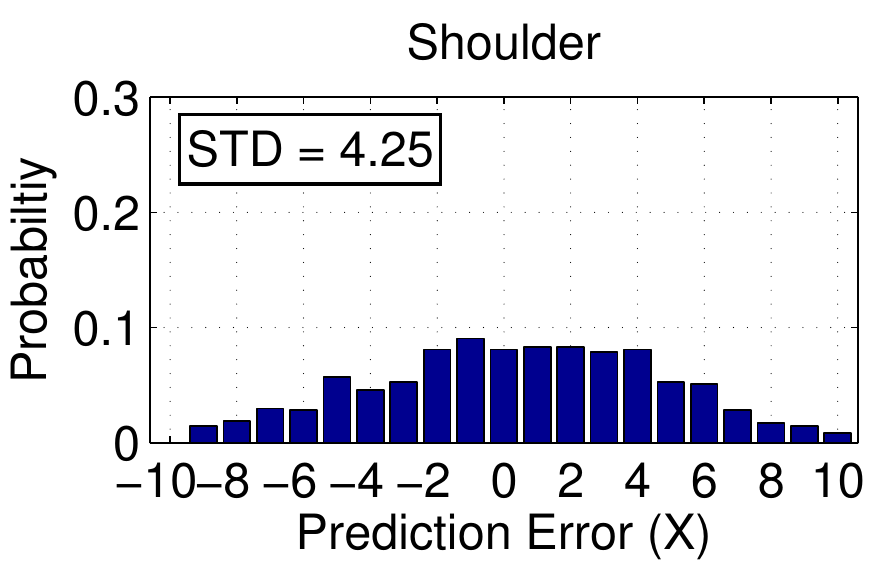}
        \caption{Coarse model only}
        \label{fig:histogram_before}
  \end{subfigure}
  \begin{subfigure}[b]{0.49\linewidth}
        \includegraphics[width=\textwidth]{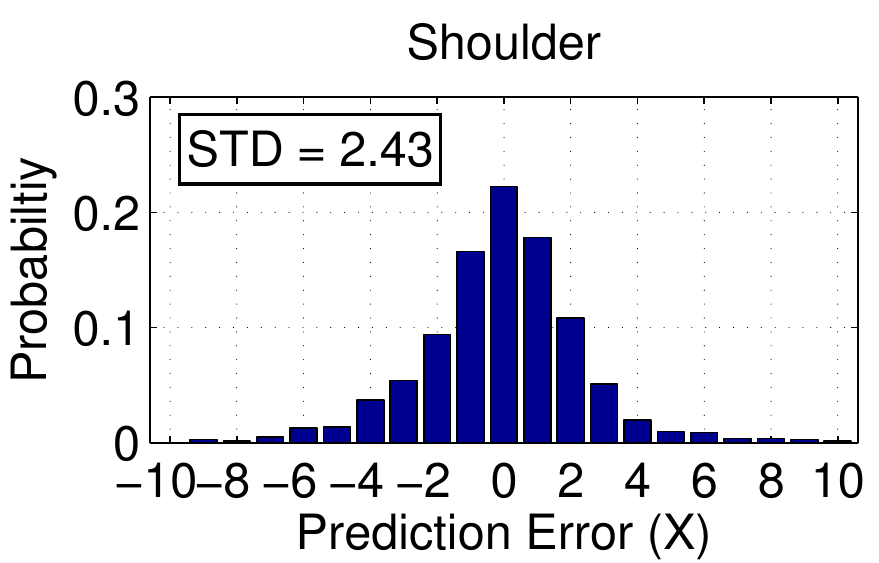}
        \caption{Cascaded model}
        \label{fig:histogram_after}
  \end{subfigure}
  \caption{Histogram of X error on FLIC test-set}
  \label{fig:histogram}
\end{figure}

The histogram of the coarse heat-map model pixel error (in the $x$ dimension) on the FLIC test-set when using an 8x internal pooling is shown in Figure~\ref{fig:histogram_before} (for the face and shoulder joints). For demonstration purposes, we quote the error in the pixel coordinates of the input image to the network (which for FLIC is $360\times240$), not the original resolution. As expected, in these coordinates there is an approximately uniform uncertainty due to quantization of the heat-map within -4 to +4 pixels. In contrast to this, the histogram of the cascaded network is shown in Figure~\ref{fig:histogram_after} and is close to the measured label noise\footnote{When calculating $\sigma$ for our model, we remove all outliers with $\text{error}>20$ and $\text{error}<-20$. These outliers represent samples where our weak spatial model chose the wrong person's joint and so do not represent an accurate indication of the spatial accuracy of our model.}.

\begin{figure}[th]
  \centering
  \begin{subfigure}[b]{0.49\linewidth}
        \includegraphics[width=\textwidth]{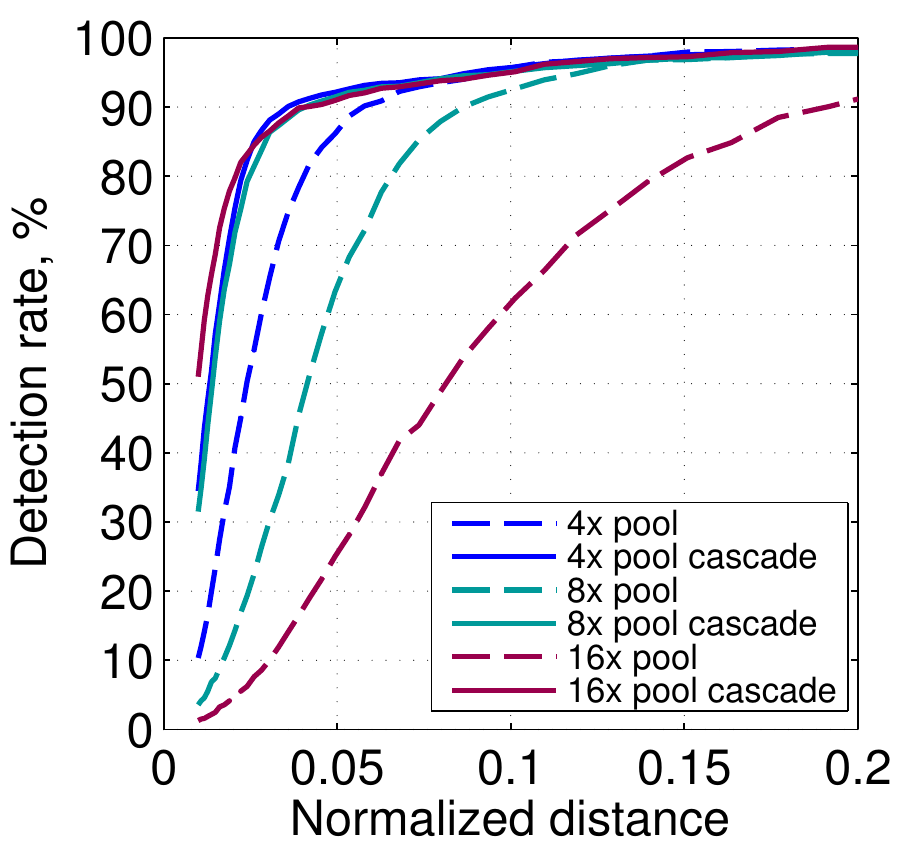}
        \caption{Face}
        \label{fig:before_after_face}
  \end{subfigure}
  \begin{subfigure}[b]{0.49\linewidth}
        \includegraphics[width=\textwidth]{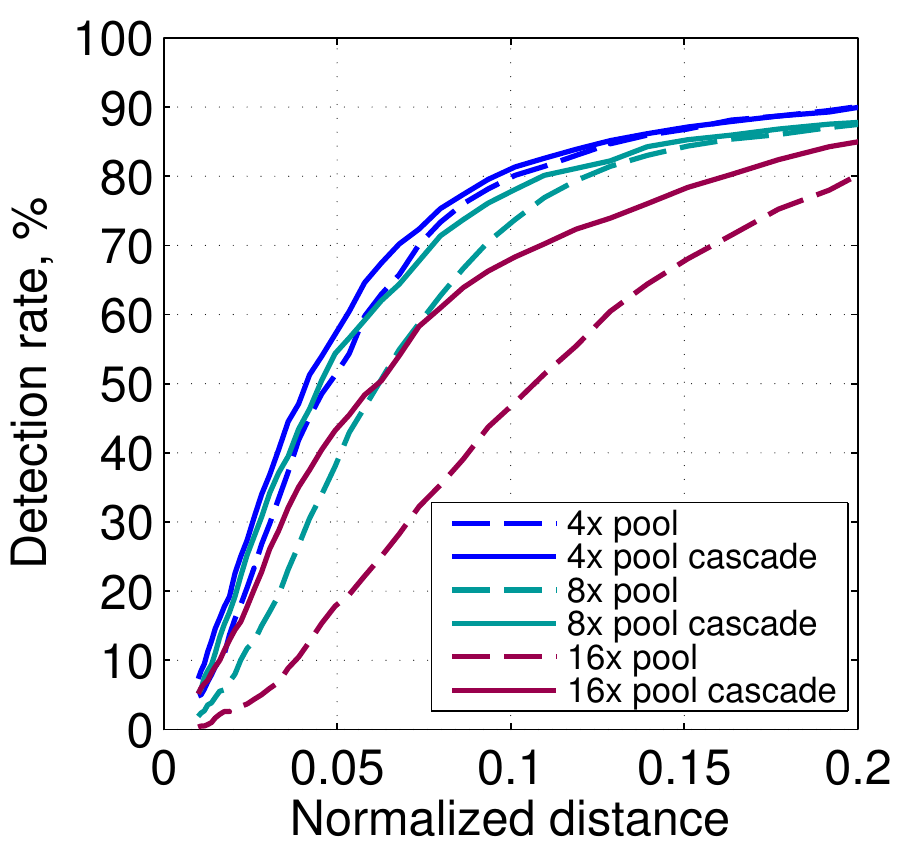}
        \caption{Wrist}
        \label{fig:before_after_wrist}
  \end{subfigure}
  \caption{Performance improvement from cascaded model}
  \label{fig:before_after}
\end{figure}

PCK performance on FLIC for face and wrist are shown in Figures~\ref{fig:before_after_face} and \ref{fig:before_after_wrist} respectively. For the face, the performance improvement is significant, especially for the $8\times$ and $16\times$ pooling part models. The FPROP time for a single image (using an Nvidia-K40 GPU) for each of our models is shown in Table~\ref{tab:times}; using the $8\times$ pooling cascaded network, we are able to perform close to the level of label noise with a significant improvement in computation time over the $4\times$ network.
\begin{table}
\begin{center}
\begin{scriptsize} 
\setlength{\tabcolsep}{0.2pt}
\begin{tabular}{ l x{1cm} x{1cm} x{1cm}}
  \hline
  \noalign{\vskip 1mm}
  
                                           & 4x pool & 8x pool & 16x pool \\
  \noalign{\vskip 1mm}
  \hline
  \noalign{\vskip 1mm}
                     Coarse-Model          & 140.0   & 74.9    & 54.7     \\
                     Fine-Model            & 17.2    & 19.3    & 15.9     \\
                     Cascade               & 157.2   & 94.2    & 70.6     \\
  \noalign{\vskip 1mm}
  \hline
\end{tabular}
\end{scriptsize}
\end{center}
\caption{Forward-Propagation time (milli seconds) for each of our FLIC trained models}
\label{tab:times}
\end{table}

The performance improvement for wrist is also significant but only for the $8\times$ and $16\times$ pooling models. Our empirical experiments suggest that wrist detection (as one of the hardest to detect joints) requires learning features with a large amount of spatial context. This is because the wrist joint undergoes larger amounts of skeletal deformation than the shoulder or face, and typically has high input variability due to clothing and wrist accessories. Therefore, with limited convolution sizes and sampling context in the fine heat-map regression network, the cascaded network does not improve wrist accuracy beyond the coarse approximation.
\begin{figure}[th]
\begin{center}
\includegraphics[width=0.85\columnwidth]{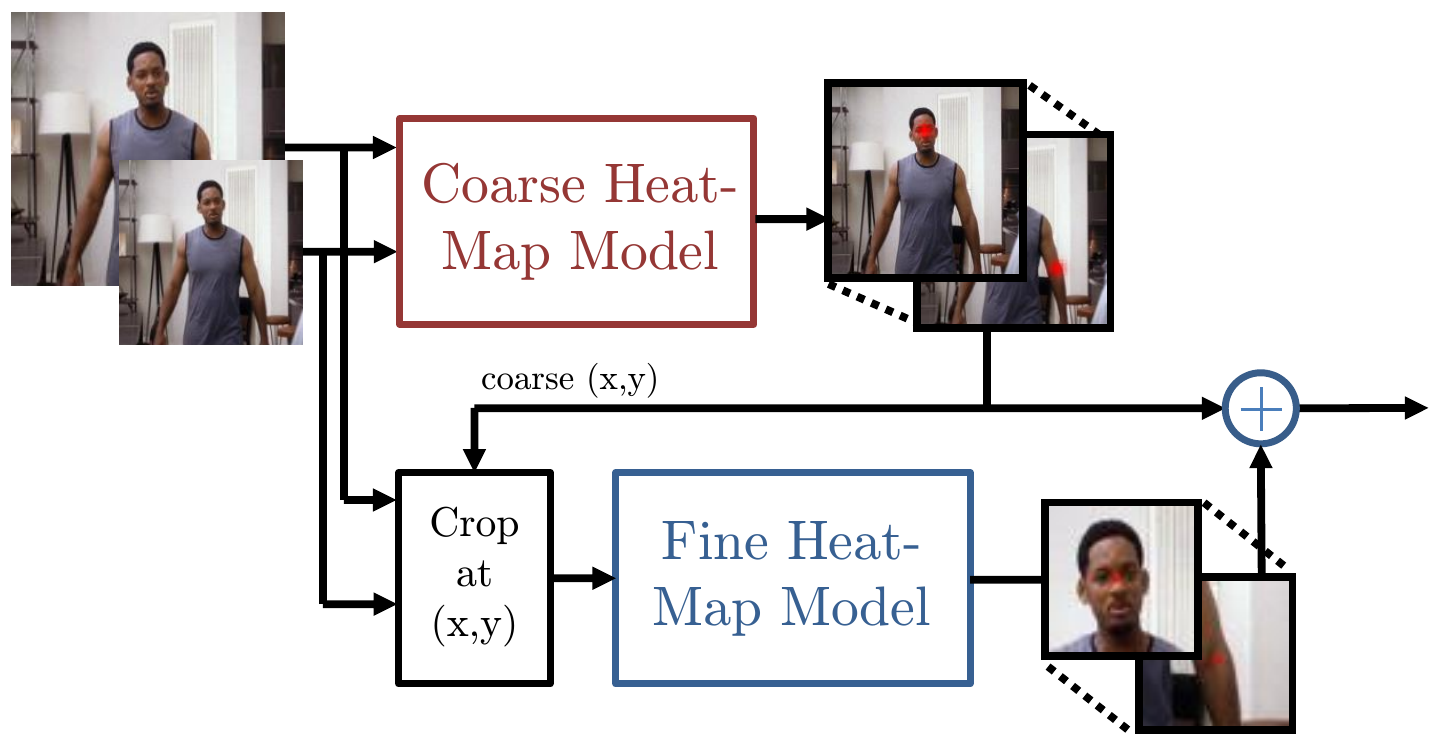}
\end{center}
\caption{Standard cascade architecture}
\label{fig:standard_cascade}
\end{figure}

\begin{figure}[th]
  \centering
  \begin{subfigure}[b]{0.49\linewidth}
        \includegraphics[width=\textwidth]{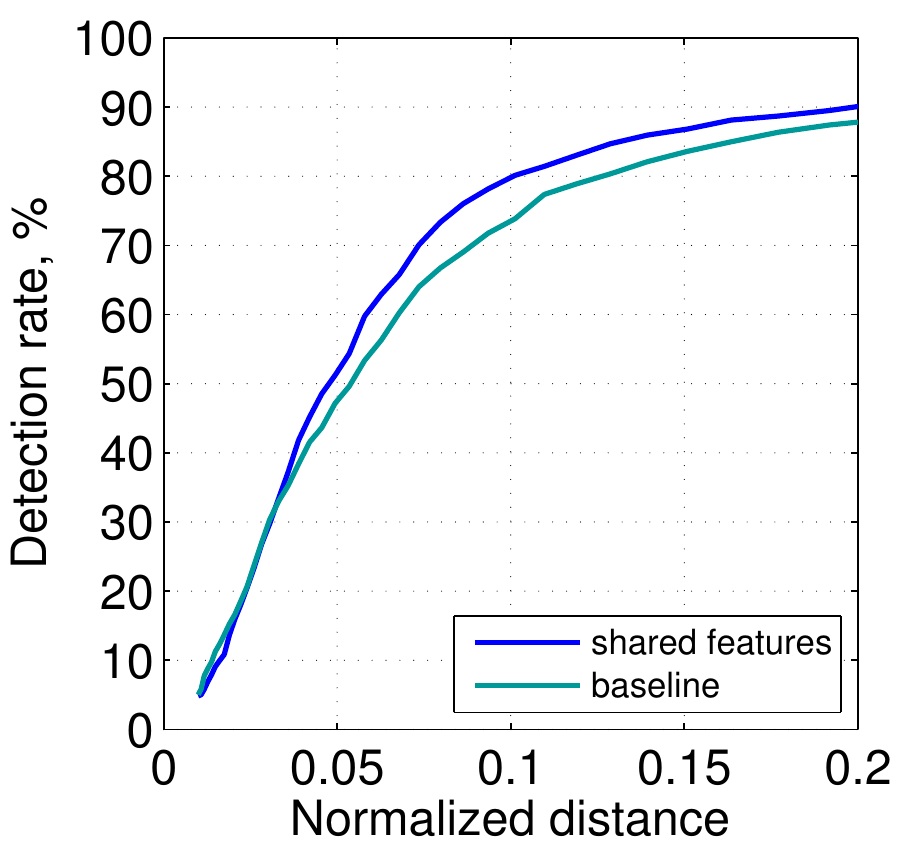}
        \caption{Ours Vs. Standard Cascade}
        \label{fig:stdreg_lwri}
  \end{subfigure}
  \begin{subfigure}[b]{0.49\linewidth}
        \includegraphics[width=\textwidth]{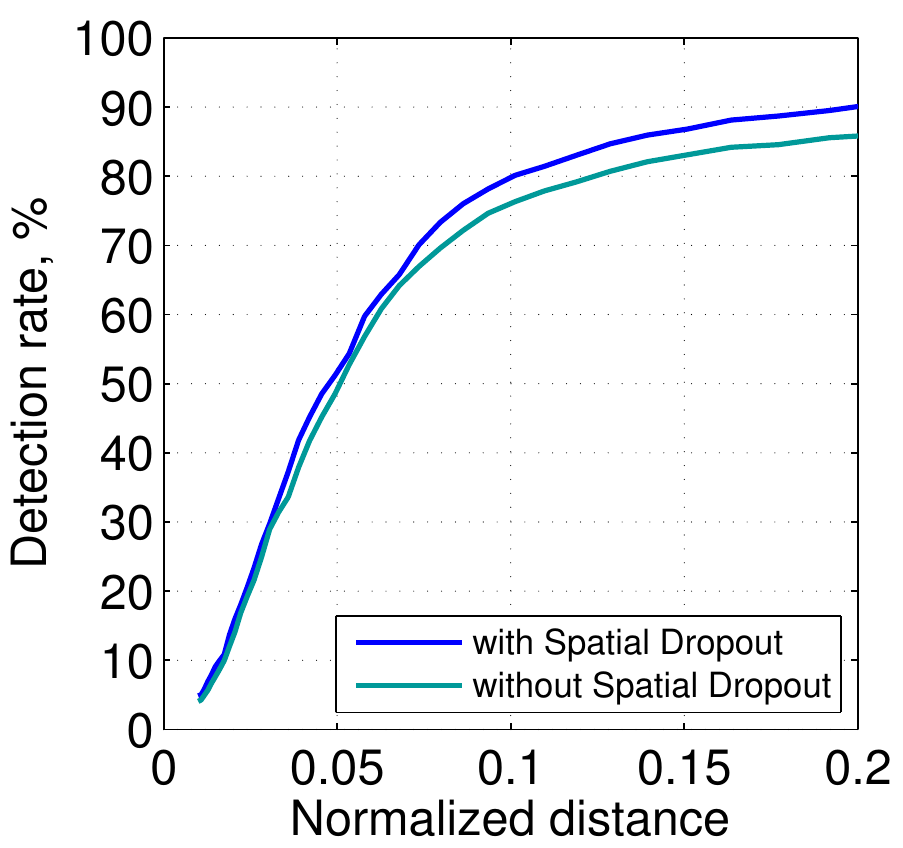}
        \caption{Impact of SpatialDropout}
        \label{fig:no_dropout_lwri}
  \end{subfigure}
  \caption{FLIC wrist performance}
  \label{fig:stdreg}
\end{figure}

To evaluate the effectiveness of the use of shared features for our cascaded network we trained a fine heat-map model (shown in Figure~\ref{fig:standard_cascade}) that takes a cropped version of the input image as it's input rather than the first and second layer convolution feature maps of our coarse heat-map model. This comparison model is a greedily-trained cascade, where the coarse and fine models are trained independently. Additionally, since the network in Figure~\ref{fig:reg_overview} has a higher capacity than the comparison model, we add an additional convolution layer such that the number of trainable parameters is the same. Figure \ref{fig:stdreg_lwri} shows that our 4x pooling network outperforms this comparison model on the wrist joint (we see similar performance gains for other joints not shown). We attribute this to the regularizing effect of joint training; the fine heat-map model term in the objective function prevents over-training of the coarse model and vice-versa.

We also show our model`s performance with and without SpatialDropout for the wrist joint in Figure~\ref{fig:no_dropout_lwri}. As expected we see significant perform gains in the high normalized distance region due to the regularizing effect of our dropout implementation and the reduction in strong heat-map outliers.

\begin{figure}[th]
  \centering
    \begin{subfigure}[b]{0.875\linewidth}
          \begin{flushright}
                  \includegraphics[trim=0.0cm 0.4cm 0.2cm 0.4cm, clip=true, width=\textwidth]{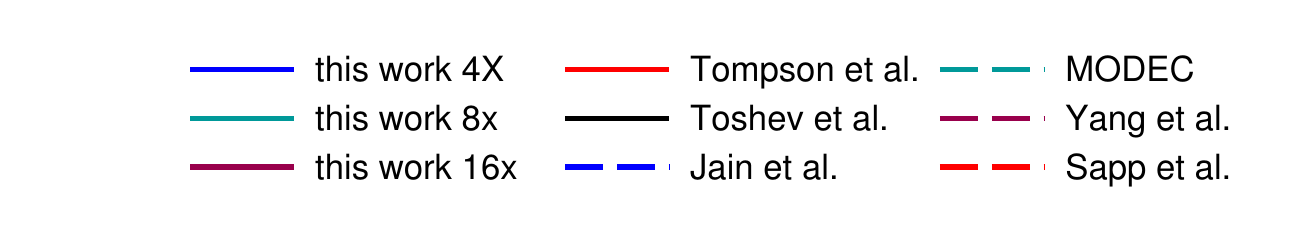}
          \end{flushright}
    \end{subfigure}
  \begin{subfigure}[b]{0.875\linewidth}
        \includegraphics[width=\textwidth]{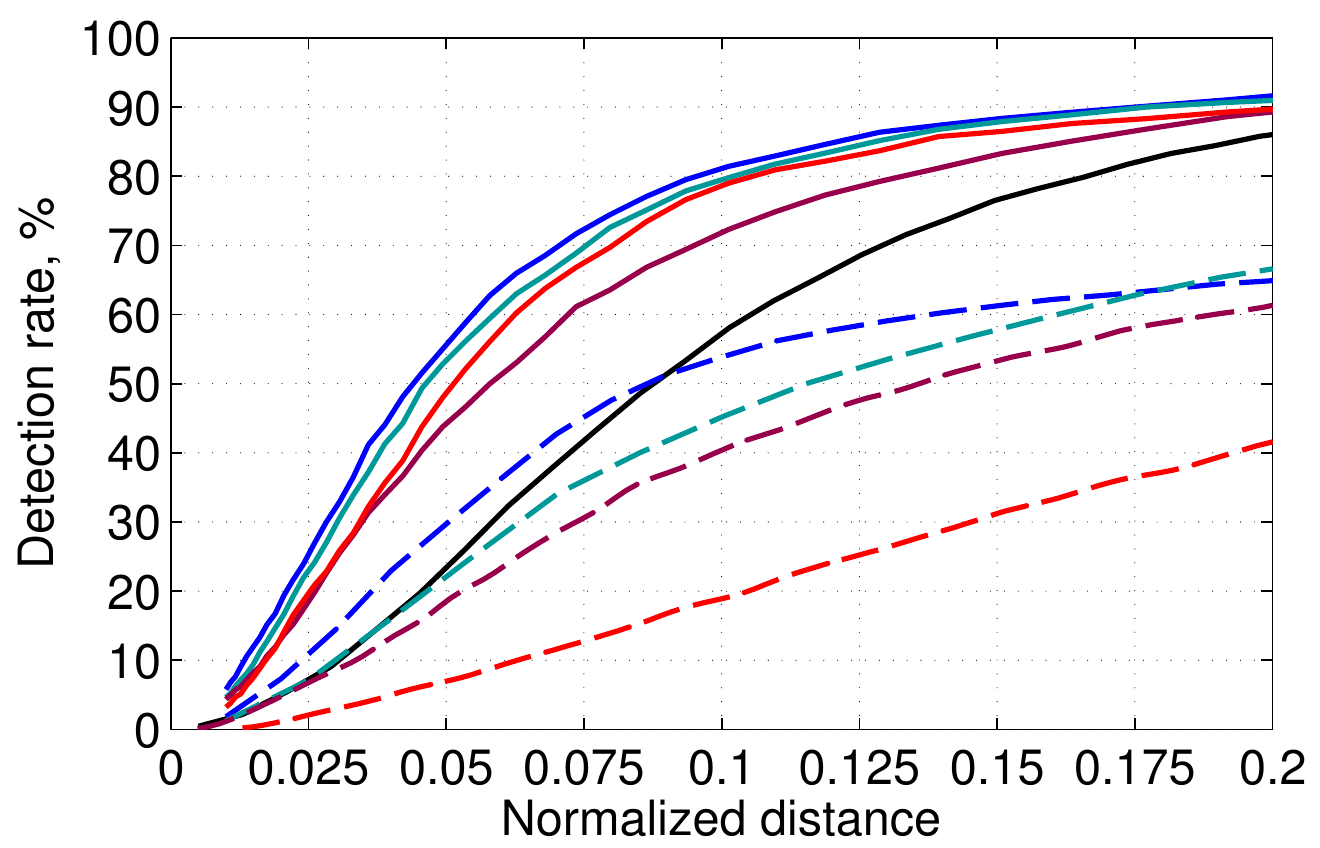}
  \end{subfigure}
  \caption{FLIC - average PCK for wrist and elbow}
  \label{fig:flic_results}
\end{figure}

Figure~\ref{fig:flic_results} compares our detector's PCK performance averaged for the wrist and elbow joints with previous work. Our model outperforms the previous state-of-the-art results by Tompson et al.~\cite{tompsonnips2014} for large distances, due to our use of \textit{SpatialDropout}. In the high precision region the cascaded network is able to out-perform all state-of-the-art by a significant margin. The PCK performance at a normalized distance of 0.05 for each joint is shown in Table~\ref{tab:flic}. 

\begin{table}
\begin{center}
\begin{scriptsize} 
\setlength{\tabcolsep}{0.2pt}
\begin{tabular}{ l x{0.7cm} x{1.3cm} x{0.85cm} x{0.75cm} }
  \hline
  \noalign{\vskip 1mm}
  
                     & Head & Shoulder & Elbow & Wrist \\
  \noalign{\vskip 1mm}
  \hline
  \noalign{\vskip 1mm}
Yang et al. & - & - & 22.6 & 15.3 \\
Sapp et al. & - & - & 6.4 & 7.9 \\
Eichner et al. & - & - & 11.1 & 5.2 \\
MODEC et al. & - & - & 28.0 & 22.3 \\
Toshev et al. & - & - & 25.2 & 26.4 \\
Jain et al. & - & 42.6 & 24.1 & 22.3 \\
Tompson et al. & 90.7 & 70.4 & 50.2 & 55.4 \\
This work 4x & \textbf{92.6} & 73.0 & \textbf{57.1} & \textbf{60.4} \\
This work 8x & 92.1 & \textbf{75.8} & 55.6 & 56.6 \\
This work 16x & 91.6 & 73.0 & 47.7 & 45.5 \\
  \noalign{\vskip 1mm}
  \hline
\end{tabular}
\end{scriptsize}
\end{center}
\caption{Comparison with prior-art on FLIC (PCK @ 0.05)}
\label{tab:flic}
\end{table}

\begin{figure}[th]
\begin{center}
\includegraphics[width=0.875\columnwidth]{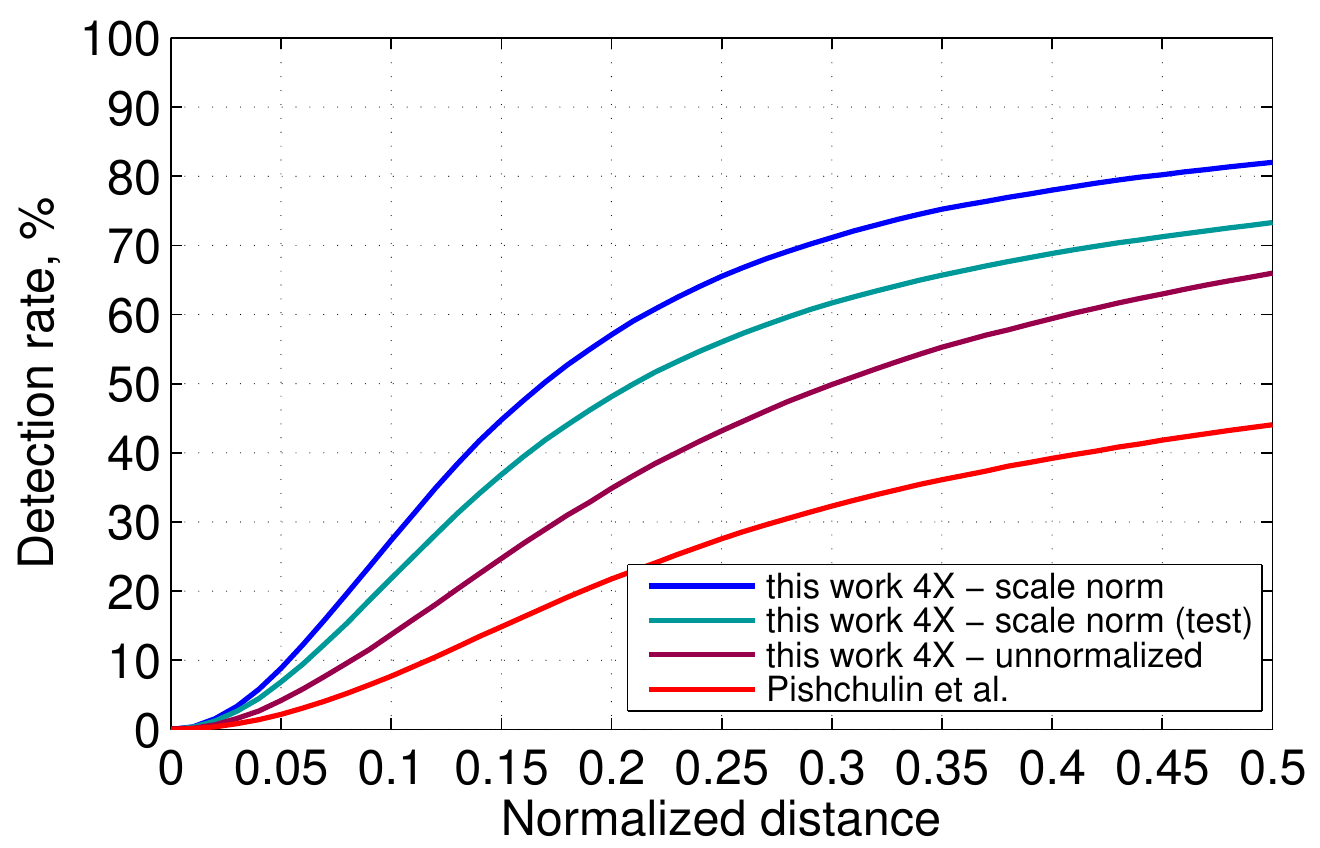}
\end{center}
   \caption{MPII - average PCKh for all joints}
\label{fig:mpi_results}
\end{figure}

Finally, Figure~\ref{fig:mpi_results} shows the PCKh performance of our model on the MPII human pose dataset. Similarity, table~\ref{tab:mpii} shows a comparison of the PCKh performance of our model and previous state-of-the-art at a normalized distance of 0.5. Our model out-performs all existing methods by a considerable margin.

\begin{table}
\begin{center}
\begin{footnotesize} 
\setlength{\tabcolsep}{0.2pt}
\resizebox{\linewidth}{!}{%
\begin{tabular}{ l x{0.6cm} x{1.2cm} x{0.75cm} x{0.65cm} x{0.7cm} x{0.6cm} x{0.8cm} x{0.75cm} x{0.6cm} }
  \hline
  \noalign{\vskip 1mm}
  
                     & Head & Shoulder & Elbow & Wrist &  Hip & Knee & Ankle & Upper Body & Full Body \\
  \noalign{\vskip 1mm}
  \hline
  \noalign{\vskip 1mm}
  Gkioxari et al.    &    - &     36.3 &  26.1 &  15.3 &    - &    - &     - &       25.9 &         - \\
  Sapp \& Taskar     &    - &     38.0 &  26.3 &  19.3 &    - &    - &     - &       27.9 &         - \\
  Yang \& Ramanan    & 73.2 &     56.2 &  41.3 &  32.1 & 36.2 & 33.2 &  34.5 &       43.2 &      44.5 \\
  Pishchulin et al.  & 74.2 &     49.0 &  40.8 &  34.1 & 36.5 & 34.4 &  35.1 &       41.3 &      44.0 \\ 
  This work - scale normalized      & \textbf{96.1} & \textbf{91.9} & \textbf{83.9} & \textbf{77.8} & \textbf{80.9} & \textbf{72.3} & \textbf{64.8} & \textbf{84.5} & \textbf{82.0} \\ 
  This work - scale normalized (test only)    & 93.5  & 87.5  & 75.5  & 67.8  & 68.3  & 60.3 & 51.7 & 77.0 & 73.3  \\ 
  This work - unnormalized   & 83.4  & 77.5  & 67.5  & 59.8  & 64.6  & 55.6 & 46.1 & 68.3 & 66.0  \\ 
  \noalign{\vskip 1mm}
  \hline
\end{tabular}}
\end{footnotesize}
\end{center}
\caption{Comparison with prior-art: MPII (PCKh @ 0.5)}
\label{tab:mpii}
\end{table}

Since the MPII dataset provides the subject scale at test-time, in standard evaluation practice the query image is scale normalized so that the average person height is constant, thus making the detection task easier. For practical applications, a query image is run through the detector at multiple scales and typically some form of non-maximum suppression is used to aggregate activations across the resultant heat-maps. An alternative is to train the ConvNet at the original query image scale (which varies widely across the test and training sets) and thus learning scale invariance in the detection stage.  This allows us to run the detector at a single scale at test time, making it more suitable for real-time applications. In Figure~\ref{fig:mpi_results} and table~\ref{tab:mpii} we show the performance of our model trained on the original dataset scale (unnormalized); we show performance of this model on both the normalized and unnormalized test set. As expected, performance is degraded as the detection problem is harder. However, surprisingly this model also out performs state-of-the-art, showing that the ConvNet is able to learn some scale invariance.

\section{Conclusion}

Though originally developed for the task of classification~\cite{LeCun1998}, Deep Convolutional Networks have been successfully applied to a multitude of other problems. In classification all variability except the object identity is suppressed. On the other hand, localization tasks such as human body pose estimation often demand a high degree of spatial precision. In this work we have shown that the precision lost due to pooling in traditional ConvNet architectures can be recovered efficiently while maintaining the computational benefits of pooling. We presented a novel cascaded architecture that combined fine and coarse scale convolutional networks, which achieved new state-of-the-art results on the FLIC~\cite{sapp13cvpr} and MPII-human-pose\cite{andriluka14cvpr} datasets.

\section{Acknowledgements}

This research was funded in part by Google and by the Office of Naval Research ONR Award N000141210327. We would also like the thank all the contributors to Torch7~\cite{torch7}, particularly Soumith Chintala, for all their hard work.


{\small
\bibliographystyle{ieee}
\bibliography{egbib}
}

\end{document}